\documentclass{article}

\usepackage{iclr2027_conference,times}

\usepackage[utf8]{inputenc} 
\usepackage[T1]{fontenc}    
\usepackage{hyperref}       
\usepackage{url}            
\usepackage{booktabs}       
\usepackage{amsfonts}       
\usepackage{nicefrac}       
\usepackage{microtype}      
\usepackage{xcolor}         
\usepackage{paralist}
\usepackage{enumitem}
\usepackage{graphicx}
\usepackage{amssymb} 
\usepackage{natbib}   

\usepackage{algorithm}
\usepackage{algpseudocode}
\usepackage{amsmath}
\usepackage{booktabs}  
\usepackage{multirow}   
\usepackage[capitalise]{cleveref}
\definecolor{linkblue}{RGB}{0,51,153}    
\definecolor{citegreen}{RGB}{0,100,35}    
\hypersetup{colorlinks=true, urlcolor=linkblue, linkcolor=linkblue, citecolor=citegreen}
\crefformat{section}{\S#2#1#3}
\Crefformat{section}{\S#2#1#3}
\crefformat{subsection}{\S#2#1#3}
\Crefformat{subsection}{\S#2#1#3}
\crefformat{subsubsection}{\S#2#1#3}
\Crefformat{subsubsection}{\S#2#1#3}
\crefformat{appendix}{\S#2#1#3}
\Crefformat{appendix}{\S#2#1#3}
\crefformat{subappendix}{\S#2#1#3}
\Crefformat{subappendix}{\S#2#1#3}
\crefformat{subsubappendix}{\S#2#1#3}
\Crefformat{subsubappendix}{\S#2#1#3}

\usepackage{listings}
\usepackage{comment}
\usepackage{footmisc}  

\usepackage{caption}
\lstdefinestyle{pythonstyle}{
    language=Python,
    basicstyle=\ttfamily\small,
    keywordstyle=\color{blue!70!black}\bfseries,
    stringstyle=\color{red!60!black},
    commentstyle=\color{gray!70!black}\itshape,
    showstringspaces=false,
    breaklines=true,
    columns=fullflexible,
}

\graphicspath{{figs/}}

\definecolor{beige}{RGB}{234,240,206}
\definecolor{silver}{RGB}{192,197,193}
\definecolor{grey}{RGB}{125,132,145}
\definecolor{lightpurple}{RGB}{87, 75, 96}
\definecolor{batchblue}{RGB}{40,100,160}
\definecolor{darkpurple}{RGB}{63, 51, 77}
\definecolor{darkred}{rgb}{0.8, 0.25, 0.33}
\definecolor{lightgray}{RGB}{245,245,245}
\definecolor{midgray}{RGB}{230,230,230}

\newcommand{\code}[1]{\texttt{#1}}
\newenvironment{tightitemize}
  {\begin{itemize}[noitemsep, topsep=0pt, leftmargin=*, labelindent=12pt]}
  {\end{itemize}}

\newcommand{\fsitem}{https://figshare.com/articles/media/Supplementary_videos_from_Modeling_Fly_Social_Behavior_with_Agent-Centric_Pose_Forecasting_/34327683}

\newcommand{\fsdoi}{https://doi.org/10.6084/m9.figshare.34327683}

\newcommand{\vidRef}{\fsitem?file=69865821}            
\newcommand{\vidRealData}{\fsitem?file=69865818}       
\newcommand{\vidShortCtx}{\fsitem?file=69865794}       
\newcommand{\vidNoDisc}{\fsitem?file=69865806}         
\newcommand{\vidDiscAll}{\fsitem?file=69865797}        
\newcommand{\vidStaticPose}{\fsitem?file=69865800}     
\newcommand{\vidNoHandPose}{\fsitem?file=69865809}     
\newcommand{\vidKeypoints}{\fsitem?file=69865812}      
\newcommand{\vidFullRef}{\fsitem?file=69865803}        
\newcommand{\vidFullRealData}{\fsitem?file=69865815}   
\newcommand{\vidRatInABox}{\fsitem?file=69865788}      

\usepackage{xspace}
\newcommand{\referenceModel}{\textsc{Reference}\xspace}
\newcommand{\shorterContextModel}{\textsc{Shorter context}\xspace}
\newcommand{\noHandcraftedPoseModel}{\textsc{No handcrafted pose}\xspace}
\newcommand{\staticLocalPoseModel}{\textsc{Static local pose}\xspace}
\newcommand{\noDiscretizationModel}{\textsc{No discretization}\xspace}
\newcommand{\discretizeAllModel}{\textsc{Discretize all}\xspace}
\newcommand{\keypointsModel}{\textsc{Keypoints}\xspace}

\newcommand{\suppvid}{S.~Video~}
\newcommand{\suppvids}{S.~Videos~}

\title{Modeling Fly Social Behavior with\\Agent-Centric Pose Forecasting}

\author{%
  Eyrun Eyjolfsdottir \& Kristin Branson\\
  HHMI Janelia Research Campus, Ashburn, VA 20147\\
  \texttt{\{eyrun.eyjolfsdottir,kristinbranson\}@gmail.com}\\
  \vspace{-.5cm}
}

\iclrpreprint
\begin{document}

\maketitle

\begin{abstract}
Understanding animal behavior at an algorithmic level -- what animals attend to, how they form internal representations, and how this maps to action -- remains a central challenge in neuroethology. We explore how well large generative models capture properties of fly social behavior. We combine transformers with agent-centric representations. Models input egocentric sensory observations and output egocentric movements, mirroring biological constraints. Social behavior emerges from agents independently sensing and responding to one another. We demonstrate that this simple model captures many properties of fly behavior, outperforming the world-centric representation and achieving SOTA on MABe 2022-fly. We systematically evaluate properties of the representation important for capturing the behavior distribution through a series of lesioning experiments. These comparisons were enabled by a new, general-purpose library for animal-centric pose forecasting\footnote{Code at \url{https://github.com/kristinbranson/AnimalPoseForecasting}\label{fn:code}}.
\end{abstract}

\section{Introduction}
\label{intro}

Large transformer-based generative models have proven exceptional at capturing the distributions of complex temporal processes, learning rich, structured knowledge about the world their training data describes \citep{li2022emergent, templeton2026scalingmonosemanticityextractinginterpretable}. Concurrently, progress in biology has led to rapid growth in dataset sizes, and supervised learning has delivered critical progress on processing this raw data — for example, video-based pose tracking now can produce large-scale measurements of behaving animals. This raises a compelling possibility: large generative models that capture the distributions of complex biological processes would give us more faithful computational descriptions than hand-crafted, simpler models. It opens the door to modeling phenomena that have so far resisted formal description, e.g.~animal behavior in more naturalistic settings, for which important events in (courtship, threat escape) are rare. Neuroscientists have made great progress by studying animals in reduced paradigms, but these designs leave out some of the complexity of animal intelligence: animals integrating information across senses, attending selectively, and balancing competing demands.

\begin{figure}
  \centering
    \includegraphics[width=.9\textwidth]{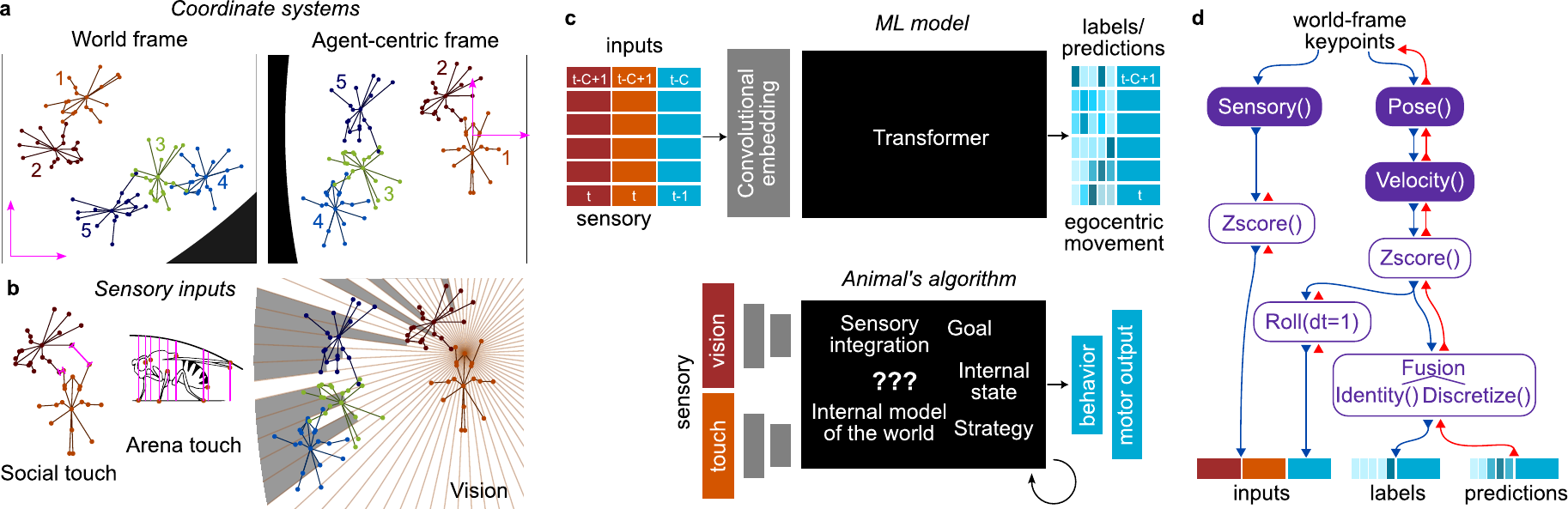}
  \caption{{\bf a.}  Unlike standard world-frame approaches, we model behavior in the agent-centric frame. {\bf b.} Hand-designed sensory features encode what each fly perceives. {\bf c.} A transformer predicts egocentric movement, paralleling the fly's sensory-to-motor mapping; multiple flies interact by sensing each other rather than centralized control. {\bf d.} Our library composes biologically (solid) and ML-motivated (outlined) transformations from world-frame keypoints to inputs and labels, and inverts the chain during rollout to produce the next step's inputs.}
  \vspace{-.2cm}
  \label{fig:overview}
\end{figure}

In this work, we address the problem of building generative models of animal behavior from pose trajectories, focusing on the social behavior of \emph{Drosophila melanogaster}, an important model organism in neuroscience, ethology, and genetics \citep{ANDERSON201418}. A challenge in adapting the success of LLMs to any new data modality is differences between the structure of language and the new modality. The temporal structure of behavior differs from that of text in important ways. Animals must react on millisecond timescales to changes in their environment, yet behavior also unfolds over much longer horizons, often in sustained bouts that resemble a textured segment more than a sequence of events (grooming, walking, sleeping). Behavior is continuous-valued, multi-dimensional, and embedded in a sensorimotor loop where actions change what the animal observes.

Beyond these challenges lies a deeper one: models must be informative about the algorithms animals use (\cref{fig:overview}c), thus we must build in constraints to model animals' internal computations, not just the output behavior. In our paradigm, each animal is treated as a separate agent that inputs sensory information and outputs movements in the body's frame (\cref{fig:overview}a). The input sensory representation is designed to limit the agent to what the animal can perceive (\cref{fig:overview}b), rather than giving it omniscient access to the full world state. Social behavior emerges from agents independently sensing and responding to other agents; there is no centralized model of the group. 

We combine agent-centric input and output representations with a standard transformer architecture (\cref{fig:overview}c). This model achieves state-of-the-art performance on the MABe 2022 fly dataset \citep{MABe2022}, which probes whether learned internal representations capture biologically meaningful properties, including which neurons were activated. We show qualitatively and quantitatively that it captures properties of fly behavior at multiple scales: fine-grained motor patterns such as coordinated leg movements while walking and behavioral motifs such as chasing and wing extension. To our knowledge, no prior generative model of fly behavior operates at the level of detailed pose in a multi-animal setting. 

We rigorously dissect the importance of a range of properties of the input and output representations by comparing a range of representation variants, including an agent-centric vs world-centric representation, deterministic regression vs discretized probabilistic prediction, predicting static pose vs velocity, and keypoints vs a kinematic-skeleton representations. We assess these model variants over a a variety of metrics that each assess different properties of how well the model captures the real behavior distribution, and find that agent-centric representation and probabilistic prediction are key. 

To build these model variants and allow exploration of future variants, we developed a library for designing, training, and simulating from these agent-centric models of animal behavior. The library is organized around composable and invertible sequences of operations (\cref{fig:overview}d) that connect parallel representations of the data, for example egocentric sensory inputs, allocentric (world-frame) pose, and egocentric movement. Operations in these sequences can be biologically motivated, like the conversions between the representations above, or ML-motivated, for example z-scoring continuous values and discretizing outputs into tokens. Invertibility is essential for autoregressive rollout, which requires inverting predicted egocentric movements back to the shared world coordinate frame, then recomputing each agent's sensory representations for the next step. No such inversion is required in LLMs, where the prediction is fed directly back as input. 

We see libraries of this kind as essential infrastructure for the emerging effort to use data-driven, predictive models as tools for scientific discovery. Mechanistic interpretability is a promising path toward this goal: by interrogating a trained model, we can extract structure in the phenomenon the model has learned to capture, including patterns that scientists have not previously articulated. Realizing this potential will require iterative work by computational and theoretical neuroscientists on what experiments to model, how to constrain models to be biologically meaningful, and how to combine them with existing theory-driven models. Whether this approach can deliver scientific insight in the messy domain of animal behavior is an open empirical question. 

Our main contributions are as follows. 
\begin{tightitemize}
\item We show that an \textbf{agent-centric} model captures many properties of fly behavior, outperforms a world-centric approach across a variety of metrics, and achieves SOTA on MABe 2022-fly. 
\item We developed a new, general-purpose library \texttt{AnimalPoseForecasting}\footref{fn:code} that allowed us to build and compare a variety of representation choices.
\item We rigorously compared different representation choices to gain insight into what is important for modeling different properties of fly behavior, and directions for future development
\end{tightitemize}

\section{Related work}
\label{sec:related}
Across autonomous driving, robotics, and neuroscience, recent work has converged on a shared problem: building generative models of agents' actions or poses, with different goals and criteria. Many works explore similar models to ours, including using transformer architectures and comparing pose representations. Our work differs in a) its focus on agent-centric representations to biologically constrain the model, b) its systematic evaluation of representation choices for fly social behavior, and c) our new agent-centric library. 

\vspace{-.5\baselineskip}

\paragraph{Prediction/forecasting.} A large body of work motivated by autonomous driving and robot navigation tries to predict where agents will go \citep{HumanTrajectoryPredictionSurvey2020, ReviewPedestrianTrajectory2022, MultiTrajPredictionSurvey2025}, with architectures including RNNs, GANs, graph networks, and transformers \citep{SocialLSTM2016, DESIRE2017, SocialGAN2018, Trajectron2019, AgentFormer2021}. Related to our approach, \citet{MotionLM} casts trajectory forecasting as language modeling over discretized motion tokens. Multi-agent interactions are handled by jointly modeling all agents with pooling, attention, or graph mechanisms, rather than restricting each agent's input to what it can sense. A related line of work forecasts full-body 3D pose, exploring representations including global/local decompositions \citep{Adeli_2021}, keypoints \citep{SoMoFormer2022} vs.\ rotations along the kinematic tree \citep{FragkiadakiLM15, Pavllo2018}, and pose vs.\ velocity \citep{martinez2017humanmotionpredictionusing}. A successful approach predicts all future time points at once in a DCT basis \citep{mao2020learningtrajectorydependencieshuman, SoMoFormer2022, rahman2023bestpractices2bodypose}, which is not naturally suited to closed-loop interaction. In both literatures, models are evaluated on short-horizon forecast accuracy, e.g.~best-of-$K$ displacement or MPJPE \citep{Human36M}, rather than on matching the distribution of real behavior.

\vspace{-.5\baselineskip}

\paragraph{Imitation learning.} Imitation learning aims to recover policies from expert demonstrations. Behavior cloning \citep{Alvinn1988, BehaviorCloningAllYouNeed2024} is equivalent to next-step trajectory prediction, and shares the same data and architectural choices as the approaches above. Reinforcement learning has been framed as sequence modeling with transformers \citep{DecisionTransformer2021, TrajectoryTransformer2021, MultiAgentRLSequence2022}, a paradigm that has scaled to vision-language-action (VLA) models that map robot camera observations to actions~\citep{RT2, OpenVLA}. These methods either act in the real world or external simulators \citep{ALE, mujoco}. Physics simulators require detailed body models and 3D kinematic data and are computationally expensive. Driving simulators use simple kinematic models \citep{Waymax2023}, but are only applicable to agents represented as rigid bounding boxes. Our library instead handles rollout through composable, invertible transformations between representations of the world, and is both lightweight and adaptable to articulated bodies and user-defined pose and sensory representations.

\vspace{-.5\baselineskip}

\paragraph{Computational models of animal behavior.} Statistical models of animal behavior such as switching linear dynamical systems segment behavior into interpretable motifs \citep{Linderman2017, Batty2019, Weinreb2024}, and generalized linear models have linked sensory cues to behavioral output \citep{Calhoun2019}. These trade expressiveness for interpretability. Another line of work aims for mechanistic realism by by mirroring animals' internal structures, from whole-body biomechanical simulations in MuJoCo \citep{flybody2025, virtualrodent2024} to connectome-constrained networks \citep{Lappalainen2024}. Our framework targets an algorithmic level of description, asking what an animal must compute given its sensory access, which complements mechanistic models of how those computations are implemented. Closest to our work, \citet{Eyrun2016} modeled fly behavior with an agent-centric RNN; we extend this to full pose trajectories and transformers, systematically compare representational choices, and evaluate with distribution-level metrics building on \citet{ImBranson}.

\section{Fly social behavior modeling}
\label{sec:fly-social-behavior}

We focus on modeling the social behavior of groups of \emph{Drosophila} \citep{flydisco}. Decades of work on fly courtship and aggression have characterized behaviors such as chasing, wing extension, and lunging, as well as some of the sensory cues like vision and taste that drive them. However, many questions about the algorithmic basis of these behaviors remain open. What do flies internally represent about their environments? Do they keep track of individual identities or social rank, and over what time scales? Do they internally represent their position in the world? How do flies choose whom to court? What strategies do they use to pursue their target while ignoring other distractors?

We base our analyses on a large video dataset of groups of $\sim 9$ socially interacting flies \citep{flydisco}, an extended version of the MABe 2022 fly dataset \citep{FlyMABe2022v3} (\cref{fig:mabe_fly_data}). In total, there are 150 videos of 1,400 flies and 72M fly frames. This dataset is enriched for courtship and aggression behaviors by activating neurons known to cause these behaviors and by manipulating the flies' social rearing conditions. Flies are thus highly responsive to their social environments. 19 keypoints describing the head, body, and limb positions have been tracked.

\section{Agent-centric model}
\label{sec:model}

Our models are based on a standard causal transformer architecture, with the main focus on choosing sensory input (\cref{sec:sensory_input}) and action output (\cref{sec:action_output}) representations of the pose trajectories. Choices here must balance biologically-motivated constraints and constraints based on properties of gradient descent and machine learning. This section describes how our framework navigates these constraints. 

\subsection{Sensory input}
\label{sec:sensory_input}

The inputs to the network are scientist-defined approximations of the sensory information available to the animal (\cref{fig:overview}b). The level of abstraction built into the input representation is a modeling decision that determines which biological questions are most addressable: any representation makes some information explicit at the expense of information that becomes hard to recover \citep{Marr1982}. We represent visual information about other flies as a depth map: the distance to the nearest fly at sampled angular directions in the focal fly's reference frame. We represent tactile information about the environment as the height of the domed ceiling at each keypoint, and tactile information about other flies as pairwise distances between body parts of the focal fly and others, sensed only when those distances are small enough for contact. These choices lie between raw sensory input and abstract task variables: they are not literal photoreceptor or mechanoreceptor responses, but quantities the animal could plausibly compute from its sensory periphery.

Each sensory modality is embedded independently with a shallow temporal and/or spatial convolutional network. The resulting embeddings are summed into a shared representation that is passed to the transformer. Temporal convolutions are used for all modalities, since sensory systems are often tuned to temporal change, e.g.~the fly visual system contains neuronal cell types that code motion direction \citep{Gruntman2018}. We include spatial convolutions only for the vision modality, whose depth map is sampled on a regular angular grid; the other modalities are measured at sparse, irregular sets of keypoints. Processing each modality separately before integration loosely parallels the organization of the fly brain, where different senses are processed in distinct neuropils before converging in higher brain centers \citep{ito2014systematic, CurrierNagel2020, Okray2023}. It also serves a practical purpose: by embedding every modality to the same dimension before summing, we balance their representational footprint during training. Without this, a high-dimensional modality (such as the visual depth map) would dominate over lower-dimensional ones (such as ceiling height). 

\subsection{Action output}
\label{sec:action_output}

The transformer's output is passed to an action-prediction head. Unlike the input representation, the output representation substantially affects training and what the model can capture, e.g.~multimodal distributions. We explore choices along three axes.

The first axis is the form of the prediction. Continuous outputs predict movement values directly and are trained with an L1 loss; discrete outputs bin each movement dimension into a vocabulary and are trained with cross-entropy. Discrete outputs can represent multimodal distributions, but introduce quantization error, and sampling each dimension independently can combine values from different modes into an incoherent movement. Continuous outputs avoid both problems, but collapse each dimension's conditional distribution to its median. Hybrid representations discretize dimensions where a median prediction would lose the most structure, such as multiple modes or rare movements.

The second axis is the temporal frame of reference: predicting the pose at each frame or its change between frames. Because behavior is strongly autocorrelated, frame-to-frame changes are concentrated near zero. For discretized outputs, velocity bins can be sized to resolve small movements finely, whereas bins over absolute pose must span all poses. However, predicting changes makes rollouts more sensitive to error accumulation, since each step's errors are integrated into all subsequent poses, and requires integrating predictions to recover the pose needed for the next step's sensory input, a compose-and-invert operation the library is built around.

The third axis is the parameterization of pose, with three options of increasing structure: (1) 2D keypoints in world coordinates, which entangle translation, orientation, and body configuration; (2) egocentric keypoints, separating global position and heading from the remaining keypoints represented in the animal's own reference frame; and (3) a skeletal decomposition inspired by character animation \citep{FragkiadakiLM15, Pavllo2018}, in which global position and heading are augmented with with joint angles and per-segment lengths, which absorb out-of-plane rotations as foreshortening. Option (3) makes articulation explicit, separating it from translation and foreshortening, at the cost of a more elaborate transformation.

Our primary model (referred to as \textsc{Reference} model in future sections, details in \cref{sec:appendix_fly_model_architecture}) predicts frame-to-frame changes: the egocentric forward, sideways, and angular velocity, and the change in the skeletal pose decomposition (option 3). We use a hybrid form, discretizing the forward, sideways, and angular velocities and the change in wing angles, and predicting all other pose changes continuously. We discretize the body velocities because we expect them to carry the main modes of behavior, such as walking versus stopping, or approaching one fly versus another (\cref{fig:multisim_walk_bout_plots}), while being close enough to independent given model inputs that sampling them separately yields coherent movements. We discretize wing-angle changes because wing movements are rare: in initial tests with continuous wing outputs, predictions collapsed to the median change of $\approx 0$, so simulated flies rarely moved their wings. We explore the effects of other choices along these three axes (\cref{sec:model_variants}).

\section{Library designed for rollout in a dynamic environment}
\label{sec:library}

As with LLMs, behavior is simulated by applying next-timepoint prediction iteratively. The structural difference is that in text generation, the prediction is fed directly as the next input: the model produces a token, and that token is appended to the context. For an agent-centric model of animals moving in a dynamic, multi-agent environment, the next step's input depends on the state of the world, which depends in turn on the actions of all agents in the previous step. Every agent's predicted movement must be inverted and integrated to world-frame keypoints. For our main model, this involves inverting the output discretization and z-scoring, then integrating the frame-to-frame velocities to get absolute pose, then inverting our chosen pose representation (\cref{fig:overview}d, red arrows) to get world-frame keypoints. Combining this for all agents produces the world state for the next timepoint. Each agent's sensory input can then be computed from the world-state, the next network input. 

We anticipate our method requiring exploration of many input and output representation choices, first to achieve good modeling quality and second as we try to interpret the model's learned computations. To make mixing and matching different sequences of operations or adding new operations in the middle of this chain straightforward, we have developed a new Python/PyTorch-based library focused on representing these operations. Our library is built around two main abstractions.

The core abstraction is the \code{Operation} class which defines both a forward \code{apply()} method for computing sensory input and action output from keypoints for training and an \code{invert()} method used during autoregressive rollout. Many operations are not strictly invertible, e.g.~discretization loses information. Both these directions may require dataset-dependent auxiliary attributes, for example the bin edges for discretization or mean and standard deviation for z-scoring. These are encapsulated into the \code{Operation} object. \code{Data} objects contain the temporal sequence of data across agents (shape \code{(\ldots, agents, T, d)}), and the list of \code{Operation}s that produced it, enabling inversion to the original or an intermediate representation:
\begin{lstlisting}[style=pythonstyle]
    raw = Data(array=kparray, name='keypoints')
    ops = [Pose(), Velocity(), Discretize(bin_edges=bin_edges)]
    labels = apply_operations(raw, ops)
    original = invert_to_named(labels,'keypoints')
\end{lstlisting}
\cref{sec:appendix_library_design} describes the library in more detail, including the general-purpose and domain-specific \code{operation}s we defined. We show the utility of this library next in exploring different input/output representation choices for fly social behavior modeling (\cref{sec:results}), as well as generalization to a new domain, RatInABox \citep{RatInABox} (\cref{sec:appendix_ratinabox}). 

\section{Results}
\label{sec:results}

Behavior is high-dimensional, multi-scale, and stochastic: a good model must reproduce fine-grained motor patterns, longer-timescale behavioral motifs, and the statistical structure of social interactions all at once, and no single scalar summarizes success across these scales. We developed a suite of complementary evaluation criteria that each probe a different facet of behavior \citep{ImBranson, MABe2022}. Like criteria designed to evaluate LLMs, our criteria assess how well the model captures the distribution of behavior rather than how well it can forecast what an animal will do in the near future. We do these by a) simulating behavior from the model autoregressively and comparing properties of simulated and real data and b) by training probes for known behaviors from learned hidden representations. We use these criteria to compare different model variants and dissect how representation choices affect performance, and to identify directions for improvement. 

\subsection{Model variants}
\label{sec:model_variants}

We trained several different models with the same neural architecture but different input and output representations. We started with a carefully designed \referenceModel model (\cref{sec:model}), then manipulated individual components to test their importance. The \referenceModel and all lesioned model variants were trained on a subset of the data consisting of male flies from videos enriched for courtship through genetically-targeted neural activation or social isolation (47 videos, 10M fly frames). Manipulated variants are the following:

\begin{tightitemize}
\item \shorterContextModel Context length reduced from 512 to 64 frames.
\item \noDiscretizationModel All output features are continuous; no discretization is applied.
\item \discretizeAllModel All output features are discretized.
\item \staticLocalPoseModel Static local pose (e.g.~wing angle) is used rather than as its velocity (e.g.~change in wing angle); global pose velocity is unchanged.
\item \noHandcraftedPoseModel Hand-crafted local pose features (e.g.~leg angles) are replaced by the egocentric keypoint $(x,y)$ coordinates, and we predict the first derivative of those. Global pose velocity is unchanged.
\item \keypointsModel The output is represented as the original $(x,y)$ keypoints in world coordinates. Instead of egocentric sensory features, we input the original $(x,y)$ keypoints of all other flies in world coordinates. Identity ordering is arbitrary but fixed. 
Note that this variant changes more than one component at once. We include it as it corresponds to the vanilla, world-centric representation, in contrast to our agent-centric approach \citep{NEURIPS2021_2fd5d41e}.
\end{tightitemize}

Implementing each variant was straightforward with our library, it only required defining a new \code{Operation} and calling it when creating the \code{Data} objects. Both training and rollout were done without further changes. \cref{fig:flow_per_variant} visualizes the operation graph for each variant.

\vspace{-.5\baselineskip}

\paragraph{Qualitative comparison of simulated and real behavior.}

\begin{figure}
    \centering
    \includegraphics[width=.8\linewidth]{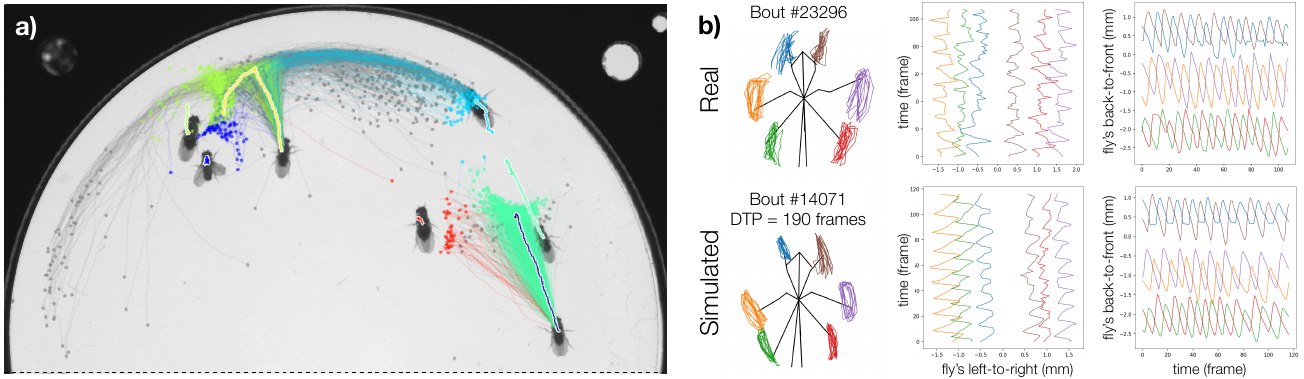}
    \caption{\textbf{a}.~1K rollouts of 128 frames each for two flies. True trajectories are white, predicted are colored by the fly they end up near, gray if they end up far from any fly. \textbf{b}.~Agent-centric leg-tip positions for a real (top) and simulated (bottom) walking fly (\referenceModel model. Both show alternating periods of swing (leg in the air) and stance (leg planted) for leg tripods. Examples from all models in \cref{fig:walk_bouts_all}.}
    \label{fig:multisim_walk_bout_plots}
\end{figure}

Videos showing simulations from randomly selected prompt sequences for all models are in \href{\fsitem}{\suppvids 1--8}. 
Simulations from the \referenceModel model (\href{\vidRef}{\suppvid 1}) qualitatively match fly behavior (\href{\vidRealData}{\suppvid 2}). Simulated flies interact with one another, with simulated flies approaching (\cref{fig:multisim_walk_bout_plots}a), chasing, and performing wing extension (courtship song) to other flies. They stay within the arena and avoid colliding with other flies even though physical collisions are not part of the simulation environment. They maintain plausible poses and their leg movements mimic tripod gait \citep{wilson1966insect} (\cref{fig:multisim_walk_bout_plots}b). Repeatedly simulating from the same prompt trajectory shows that the model's distribution is multimodal: the rollouts form distinct clusters of trajectories (\cref{fig:multisim_walk_bout_plots}a). Comparing model variants to the reference, we only see qualitative effects of output discretization: not discretizing any outputs produces mostly stationary flies, even when real flies approach and eventually walk through them (\suppvids \href{\vidNoDisc}{4}, \href{\vidKeypoints}{8}), while independently discretizing all outputs (\href{\vidDiscAll}{\suppvid 5}) results in non-realistic poses and flies that escape the arena. 

\vspace{-.5\baselineskip}
 
\paragraph{Feature distribution comparison.}

\begin{figure}[htb]
    \centering
    \includegraphics[width=1.0\linewidth]{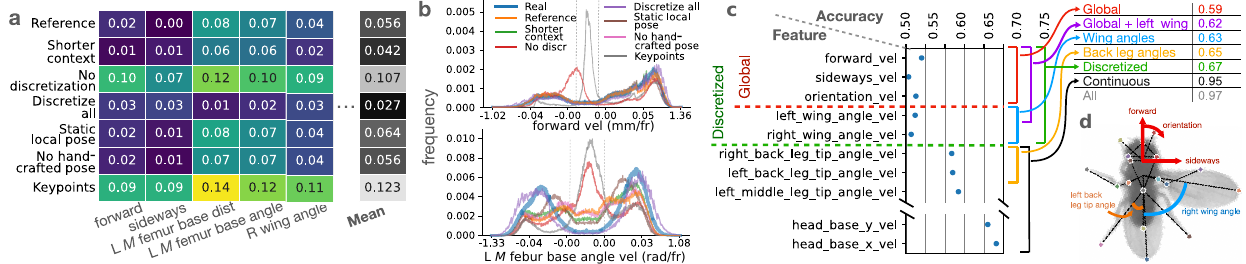}        
    \caption{\textbf{a}.~Wasserstein distance between distributions of real and simulated flies for velocity features across model variants. Full table and position/pose distances in \cref{fig:wasser_full}. Distribution distance as we simulate for longer is shown in \cref{fig:feat_per_dtp}. \textbf{b}.~Histograms of two selected pose velocity features for real and simulated flies. Thin vertical lines indicate the linear portion of the symlog binning transform. \textbf{c.} Real-vs-simulated discrimination accuracy using individual and subsets of per-frame pose-velocity features (full results in \cref{fig:discr_per_feat_all}). \textbf{d}. Illustration of example features: global velocity (red), selected static pose features.}
    \label{fig:wasser}
    \label{fig:discr_per_feat}
    
 \end{figure}

We compare the distribution of one-dimensional projections of the real and simulated data via the Wasserstein distance (\cref{fig:wasser}a, \cref{sec:wasserstein_details}). We chose 58 features that are interpretable, biologically meaningful quantities that biologists typically use when analyzing animal behavior data: the agent-centric velocity (forward, sideways, angular), the world-centric position (x, y, orientation), and hand-crafted agent-centric pose features and their velocities (\cref{sec:appendix_pose_operation}). The reference model matches the real data distribution faithfully, with an average distance of 0.005 for pose features and 0.056 for velocity features. The analysis for velocity features shows the importance of discretizing the output features: models that do not discretize output features are much worse, while the model that discretizes all features performs best (\cref{fig:wasser}a). Examining the distributions (\cref{fig:wasser}b), we see that real data is multimodal, and that do not discretize any features primarily output zero velocity. This is largely improved but not completely rescued by discretizing some output features: for continuous output features, the multi-modality is captured, but there is still a small increase in the bias for outputing zero velocity.

\vspace{-.5\baselineskip}

\paragraph{Real-vs-simulated discriminator.}

\begin{figure}
    \centering
    \includegraphics[width=1.0\linewidth]{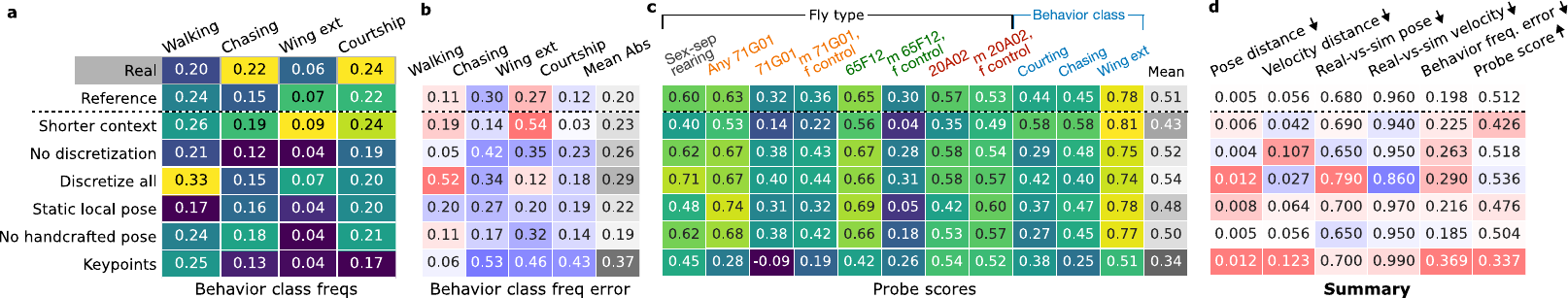}
    \vspace{-.4cm}
    \caption{\textbf{a}.~Fraction of time real and simulated flies perform each behavior across model variants. \textbf{b}.~Error ($|\text{real}-\text{sim}|/\text{real}$) between real and simulated behavior frequencies. In (b,d), red indicates performance worse than \referenceModel, blue better. \textbf{c}. Matthew's correlation coefficient (1=perfect, 0=chance) of probes from learned hidden representation to MABe 2022 categories. \textbf{d}.~Average per type of all metrics presented.}
    \vspace{-.4cm}
    \label{fig:summary_tables}
\end{figure}

Behavior is high-dimensional, and one-dimensional feature distribution comparisons do not capture correlations between features, for example coordination between legs. To this end, we train discriminators (\cref{sec:appendix_discriminator_details}) to
distinguish real from simulated data. 
Discriminators for all models perform better than chance (\cref{fig:discr_all,fig:summary_tables}), indicating that the simulated data are distinguishable from real. Surprisingly, for all models, simulated data are more distinguishable based on velocity than pose, even those that are trained to predict velocity. Here, the best-performing model discretizes all outputs independently, indicating that continuous outputs may be to blame for this discriminability. To understand this result better, we train discriminators using individual and groups of features for our reference model (\cref{fig:wasser}b). Comparing different subsets of features, we see that some of this discriminability can be explained by the independent sampling of discretized features that are correlated, but continuous features account for a much larger portion.

\vspace{-.5\baselineskip}

\paragraph{Behavior class frequencies.}

We asked whether simulated flies reproduce the same behavior patterns as real flies with the same frequency. Using real fly data, we trained classifiers for courtship behaviors (chasing, unilateral wing extension, and the superclass courting) \citep{kabra2013jaaba} and handcrafted a walking classifier. We then compared the fraction of time real vs simulated flies spend performing each behavior (\cref{fig:summary_tables}a-b). Simulated flies from our reference model perform all behaviors, despite some being relatively rare in real data. For example, real flies in our dataset perform wing extension only 6\% of the time, and simulated flies perform this behavior 7\% of the time. Comparing different model variants, we see that models that produce qualitatively convincing simulations perform similarly in terms of these more complex behavior patterns. The model that uses the world-centric (\keypointsModel) instead of the agent-centric representation is by far the worst-performing model by this metric -- 50\% worse than the directly comparable \noDiscretizationModel model. 

\vspace{-.5\baselineskip}

\paragraph{Probing model internals.}

The MABe 2022 challenge was designed to test unsupervised behavior representation learning -- given just trajectories, could a behaviorally meaningful representation of behavior be learned \citep{MABe2022}? To measure this, it evaluates how well simple linear classifiers can be trained to predict known, behaviorally relevant labels from the learned representation. These include fly-type labels that are known to influence social behavior -- which neuronal cell types are activated or behavioral conditioning, as well as frame-level, expert-annotated behavior category labels. We use these labels to test whether our model variants implicitly learned to capture these higher-level labels in their internal representations (\cref{fig:summary_tables}c). We trained linear probes on a middle transformer layer representation ($L=6/10$, $d=2048$). 

We compared our model variants trained on the subset of data enriched for courtship behaviors. By far the worst-performing model is that with the world-centric representation (\keypointsModel), 32\% worse than the directly comparable model (\noDiscretizationModel). Surprisingly, variants with relatively poor simulation quality (\discretizeAllModel, \noDiscretizationModel) performed well according to this metric, slightly better than our reference model, emphasizing that our different metrics measure different properties of the model, while the shorter context window results in worse performance. 

\vspace{-.5\baselineskip}

\paragraph{Key takeaways.}
\cref{fig:summary_tables}d summarizes all quantitative metrics discussed above across model variants.
Overall, our reference model \textbf{accurately fits the distribution of real fly behavior across many metrics}. Simulations from the model closely resemble real fly behavior (\href{\vidRef}{\suppvid 1}). Legs move in realistic gait patterns while walking (\cref{fig:multisim_walk_bout_plots}b). Flies react appropriately to their environment, turning at arena boundaries to stay within the confines of the arena, and approaching/chasing other flies while avoiding collisions. The model captures modes of the real distribution (\cref{fig:multisim_walk_bout_plots}a) across many projections (\cref{fig:wasser}a), including producing complex social behavior patterns like chasing and wing extension that occur somewhat rarely in the data. Its biggest weakness is the real-vs-sim discriminator, which we investigate to understand which properties contribute to discriminability. 

By comparing different model variants, we are able to attribute what each modeling component contributes. \textbf{In almost every metric, the \keypointsModel model, which does not use an agent-centric representation, performs worst} (\cref{fig:summary_tables}d). This is not simply a byproduct of its fully continuous output, as the \noDiscretizationModel strongly outperforms \keypointsModel despite also lacking any discretized outputs. Most striking is the poor performance of the \keypointsModel variant on the probe task (\cref{fig:summary_tables}c). This shows that the \textbf{agent-centric representation} is most important for accurate behavior modeling. 

We find that \textbf{discretized probabilistic predictions are needed to avoid the pitfall of forecasting no movement}. It is common for flies to sit still for long periods of time, and a no-movement predictor will achieve a decent next-frame prediction score. In addition, behavior is multimodal/stochastic (\cref{fig:wasser}), thus continuous, deterministic regression models end up hedging their bets and predicting no movement. 
Even hybrid models that discretize some features show a slight bias toward zero movement for continuous features, suggesting that discretizing {\em all} output features would be useful. However, the naive way of doing this, independently discretizing each feature and independently sampling from them, leads to simulations that degrade with distance from the prompt (\cref{fig:feat_per_dtp}). This suggests improvements may come from methods that quantize the full output dimensionality like Vector-Quantized VAE \citep{zhang2023t2mgptgeneratinghumanmotion}, or methods that sample each output dimension iteratively \citep{TrajectoryTransformer2021}, or other methods for modeling continuous distributions \citep{tevet2022humanmotiondiffusionmodel}. 

Finally, we note that the \referenceModel and \noHandcraftedPoseModel models perform similarly, suggesting that it is \textbf{not important to decompose local pose into meaningful angles}, but instead keypoint coordinate velocities are sufficient. Many human pose forecasting approaches represent pose based on joint rotations along the kinematic skeleton \citep{FragkiadakiLM15, Pavllo2018}; this angle representation may not be important for success. 

\subsection{Full dataset model}
\label{sec:full_model}

\begin{figure}
    \centering
    \includegraphics[width=0.6\linewidth]{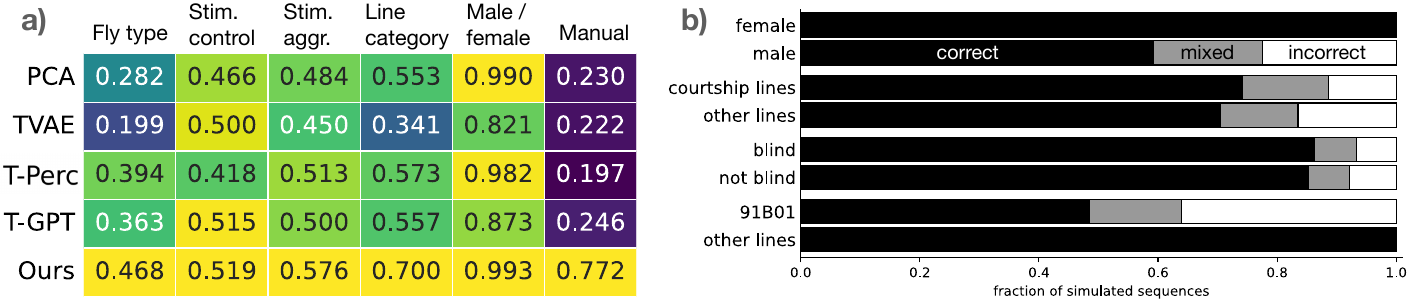}
    \caption{\textbf{a}.~F1 scores for the MABe 2022 fly challenge. "Ours" is the middle transformer layer of our \referenceModel trained on the full dataset. Performance of other methods as reported in \cite{MABe2022}. \textbf{b}.~Prompting accuracy. For each prompt fly type (bars), fraction of rollouts classified as the correct fly type.}
    \label{fig:mabe_comparison}
\end{figure}

To compare our method to baseline and published methods, we trained our \referenceModel model on the full MABe 2022 dataset \citep{MABe2022}, and tested on all 50 probe categories. On all category types, our method outperformed previous methods by a wide margin (\cref{fig:mabe_comparison}a), up to 3x the F1 score for "Manual", the average of the six expert-annotated behavior categories. We note the important caveat that our model was trained on the augmented unsupervised dataset \citep{FlyMABe2022v3}. 

\href{\vidFullRef}{\suppvid 9} show flies simulated from randomly selected prompt sequences, with the corresponding real sequences in \href{\vidFullRealData}{\suppvid 10}. As with the smaller model trained only on male flies from videos enriched for courtship, we see that it produces behavior that is qualitatively similar to real fly behavior. Here, we are simulating all flies from a wider range of fly types, and noticed that simulated behaviors differ greatly across sequences and between males and females. We asked whether prompting with one fly type produced simulations from the same fly type, as classified by the probe classifiers described above, and see that this is indeed the case (\cref{fig:mabe_comparison}b, \cref{sec:full_model}). 

\section{Discussion}
\label{conclusion}

In this work, we evaluated how well standard transformers with agent-centric sensory input and action outputs can be trained to model fly social behavior. The agent-centric framework more closely parallels the information flow in real animals, and multi-agent behavior is produced by individual agents sensing and reacting to each other. We show that our model captures properties of fly behavior at multiple scales — from fine-grained motor patterns to recognizable behavioral motifs to biologically meaningful differences across experimental conditions. Our model obtains SOTA results on the MABe 2022 fly behavior modeling challenge. Through a sequence of lesioning experiments, we show that the agent-centric representation is critical to this success, and that a world-frame model performs poorly. We also show that probabilistic prediction is critical to success, as the distribution of behaviors is multimodal and some ethologically important behaviors are rare. 

Transformers are common in trajectory and pose forecasting, while agent-centric representations are common in neuroscience and reinforcement learning. The combination of these two approaches is less common, and requires a complex sequence of operations to transform from network outputs to next-timepoint inputs during autoregressive rollouts. We have released a general-purpose library, \texttt{AnimalPoseForecasting}, that handles the representational machinery necessary. Modeling animal behavior at this level has a large design space, and the best choice is dependent on the domain and the scientific question under consideration. The library is designed to make this space explorable and accessible to the neuroscience community. 

The ultimate goal of this work is to use such models as instruments for scientific discovery, interrogating their internal structure to reveal computations the animal itself implements. This requires 1) models good enough to be worth interpreting and 2) methods for interpreting them. Our work addresses the first, providing trained models of fly social behavior, empirical evidence to guide representational choices, and a library that lets others explore alternative representations and apply them to new domains. We hope these resources will enable the second: developing methods that turn predictive models of behavior into testable hypotheses about neural computation.

\subsection*{AI use statement}

\begin{tightitemize}
\item We used generative AI to help write the text; we wrote the initial draft, LLMs provided language suggestions, and we incorporated these. 
\item We used generative AI to write scripts to help parallelize data processing across our cluster, such as running the simulations. We manually verified that the way the tasks were split for distribution was sound.
\item We used generative AI to help convert MATLAB-based JAABA to Python-based code that could run on simulations. As a sanity check, we verified that intermediate features and final classifier outputs were identical used a small dataset that was the same across all runs and verified that the classifications were unchanged between those different runs. We then manually extracted the results of that to compare behavior frequencies between real and sim.
\end{tightitemize}

We did not use generative AI for the following required disclosure tasks: Generate synthetic data sets, implement methods, support qualitative and thematic data analysis, clean and reformat dataset, help develop theoretical models or conceptual frameworks, 
formulate mathematical claims, provide critical ingredients for proving mathematical claims, 
assist in the writing of proofs, propose or refine hypotheses, design or provide feedback on research  methodology or experiments, assist with translation, interpret results. 

We did not use generative AI for the following recommended disclosure tasks: Formulate questions for surveys or interviews, suggest experimental parameters, creation of artifacts, draft parts of a research paper, transcribe recordings of research material, discover research topics or identify gaps, brainstorming, format references, suggest a structure for a research paper, 
propose a title or keywords for a research paper.

{
\small
\bibliographystyle{unsrtnat}
\bibliography{main}

\begin{thebibliography}{54}
\providecommand{\natexlab}[1]{#1}
\providecommand{\url}[1]{\texttt{#1}}
\expandafter\ifx\csname urlstyle\endcsname\relax
  \providecommand{\doi}[1]{doi: #1}\else
  \providecommand{\doi}{doi: \begingroup \urlstyle{rm}\Url}\fi

\bibitem[Li et~al.(2022)Li, Hopkins, Bau, Vi{\'e}gas, Pfister, and Wattenberg]{li2022emergent}
Kenneth Li, Aspen~K Hopkins, David Bau, Fernanda Vi{\'e}gas, Hanspeter Pfister, and Martin Wattenberg.
\newblock Emergent world representations: Exploring a sequence model trained on a synthetic task.
\newblock \emph{arXiv preprint arXiv:2210.13382}, 2022.

\bibitem[Templeton et~al.(2026)Templeton, Conerly, Marcus, Lindsey, Bricken, Chen, Pearce, Citro, Ameisen, Jones, Cunningham, Turner, McDougall, MacDiarmid, Tamkin, Durmus, Hume, Mosconi, Freeman, Sumers, Rees, Batson, Jermyn, Carter, Olah, and Henighan]{templeton2026scalingmonosemanticityextractinginterpretable}
Adly Templeton, Tom Conerly, Jonathan Marcus, Jack Lindsey, Trenton Bricken, Brian Chen, Adam Pearce, Craig Citro, Emmanuel Ameisen, Andy Jones, Hoagy Cunningham, Nicholas~L Turner, Callum McDougall, Monte MacDiarmid, Alex Tamkin, Esin Durmus, Tristan Hume, Francesco Mosconi, C.~Daniel Freeman, Theodore~R. Sumers, Edward Rees, Joshua Batson, Adam Jermyn, Shan Carter, Chris Olah, and Tom Henighan.
\newblock Scaling monosemanticity: Extracting interpretable features from claude 3 sonnet, 2026.
\newblock URL \url{https://arxiv.org/abs/2605.29358}.

\bibitem[Anderson and Perona(2014)]{ANDERSON201418}
David J. Anderson and Pietro Perona.
\newblock Toward a science of computational ethology.
\newblock \emph{Neuron}, 84\penalty0 (1):\penalty0 18--31, 2014.
\newblock ISSN 0896-6273.
\newblock \doi{https://doi.org/10.1016/j.neuron.2014.09.005}.
\newblock URL \url{https://www.sciencedirect.com/science/article/pii/S0896627314007934}.

\bibitem[Sun et~al.(2023)Sun, Marks, Ulmer, Chakraborty, Geuther, Hayes, Jia, Kumar, Oleszko, Partridge, Peelman, Robie, Schretter, Sheppard, Sun, Uttarwar, Wagner, Werner, Parker, Perona, Yue, Branson, and Kennedy]{MABe2022}
Jennifer~J. Sun, Markus Marks, Andrew Ulmer, Dipam Chakraborty, Brian Geuther, Edward Hayes, Heng Jia, Vivek Kumar, Sebastian Oleszko, Zachary Partridge, Milan Peelman, Alice Robie, Catherine~E. Schretter, Keith Sheppard, Chao Sun, Param Uttarwar, Julian~M. Wagner, Eric Werner, Joseph Parker, Pietro Perona, Yisong Yue, Kristin Branson, and Ann Kennedy.
\newblock {MABe22}: A multi-species multi-task benchmark for learned representations of behavior, 2023.
\newblock URL \url{https://arxiv.org/abs/2207.10553}.

\bibitem[Rudenko et~al.(2020)Rudenko, Palmieri, Herman, Kitani, Gavrila, and Arras]{HumanTrajectoryPredictionSurvey2020}
Andrey Rudenko, Luigi Palmieri, Michael Herman, Kris~M Kitani, Dariu~M Gavrila, and Kai~O Arras.
\newblock Human motion trajectory prediction: a survey.
\newblock \emph{The International Journal of Robotics Research}, 39\penalty0 (8):\penalty0 895–935, June 2020.
\newblock ISSN 1741-3176.
\newblock \doi{10.1177/0278364920917446}.
\newblock URL \url{http://dx.doi.org/10.1177/0278364920917446}.

\bibitem[Korbmacher and Tordeux(2022)]{ReviewPedestrianTrajectory2022}
Raphael Korbmacher and Antoine Tordeux.
\newblock Review of pedestrian trajectory prediction methods: Comparing deep learning and knowledge-based approaches, 2022.
\newblock URL \url{https://arxiv.org/abs/2111.06740}.

\bibitem[Huang et~al.(2025)Huang, Xue, Pagnucco, Salim, and Song]{MultiTrajPredictionSurvey2025}
Renhao Huang, Hao Xue, Maurice Pagnucco, Flora~D. Salim, and Yang Song.
\newblock Vision-based multi-future trajectory prediction: A survey.
\newblock \emph{IEEE Transactions on Neural Networks and Learning Systems}, 36\penalty0 (8):\penalty0 13691–13708, August 2025.
\newblock ISSN 2162-2388.
\newblock \doi{10.1109/tnnls.2025.3550350}.
\newblock URL \url{http://dx.doi.org/10.1109/TNNLS.2025.3550350}.

\bibitem[Alahi et~al.(2016)Alahi, Goel, Ramanathan, Robicquet, Fei-Fei, and Savarese]{SocialLSTM2016}
Alexandre Alahi, Kratarth Goel, Vignesh Ramanathan, Alexandre Robicquet, Li~Fei-Fei, and Silvio Savarese.
\newblock Social {LSTM}: Human trajectory prediction in crowded spaces.
\newblock In \emph{Proceedings of the IEEE conference on computer vision and pattern recognition}, pages 961--971, 2016.

\bibitem[Lee et~al.(2017)Lee, Choi, Vernaza, Choy, Torr, and Chandraker]{DESIRE2017}
Namhoon Lee, Wongun Choi, Paul Vernaza, Christopher~B. Choy, Philip H.~S. Torr, and Manmohan Chandraker.
\newblock Desire: Distant future prediction in dynamic scenes with interacting agents, 2017.
\newblock URL \url{https://arxiv.org/abs/1704.04394}.

\bibitem[Gupta et~al.(2018)Gupta, Johnson, Fei-Fei, Savarese, and Alahi]{SocialGAN2018}
Agrim Gupta, Justin Johnson, Li~Fei-Fei, Silvio Savarese, and Alexandre Alahi.
\newblock Social {GAN}: Socially acceptable trajectories with generative adversarial networks.
\newblock In \emph{Proceedings of the IEEE conference on computer vision and pattern recognition}, pages 2255--2264, 2018.

\bibitem[Ivanovic and Pavone(2019)]{Trajectron2019}
Boris Ivanovic and Marco Pavone.
\newblock The trajectron: Probabilistic multi-agent trajectory modeling with dynamic spatiotemporal graphs, 2019.
\newblock URL \url{https://arxiv.org/abs/1810.05993}.

\bibitem[Yuan et~al.(2021)Yuan, Weng, Ou, and Kitani]{AgentFormer2021}
Ye~Yuan, Xinshuo Weng, Yanglan Ou, and Kris Kitani.
\newblock Agentformer: Agent-aware transformers for socio-temporal multi-agent forecasting, 2021.
\newblock URL \url{https://arxiv.org/abs/2103.14023}.

\bibitem[Seff et~al.(2023)Seff, Cera, Chen, Ng, Zhou, Nayakanti, Refaat, Al-Rfou, and Sapp]{MotionLM}
Ari Seff, Brian Cera, Dian Chen, Mason Ng, Aurick Zhou, Nigamaa Nayakanti, Khaled~S. Refaat, Rami Al-Rfou, and Benjamin Sapp.
\newblock {MotionLM}: Multi-agent motion forecasting as language modeling.
\newblock In \emph{Proceedings of the IEEE/CVF International Conference on Computer Vision (ICCV)}, pages 8579--8590, 2023.
\newblock \doi{10.1109/ICCV51070.2023.00788}.

\bibitem[Adeli et~al.(2021)Adeli, Ehsanpour, Reid, Niebles, Savarese, Adeli, and Rezatofighi]{Adeli_2021}
Vida Adeli, Mahsa Ehsanpour, Ian Reid, Juan~Carlos Niebles, Silvio Savarese, Ehsan Adeli, and Hamid Rezatofighi.
\newblock Tripod: Human trajectory and pose dynamics forecasting in the wild.
\newblock In \emph{2021 IEEE/CVF International Conference on Computer Vision (ICCV)}, page 13370–13380. IEEE, October 2021.
\newblock \doi{10.1109/iccv48922.2021.01314}.
\newblock URL \url{http://dx.doi.org/10.1109/ICCV48922.2021.01314}.

\bibitem[Vendrow et~al.(2022)Vendrow, Kumar, Adeli, and Rezatofighi]{SoMoFormer2022}
Edward Vendrow, Satyajit Kumar, Ehsan Adeli, and Hamid Rezatofighi.
\newblock Somoformer: Multi-person pose forecasting with transformers, 2022.
\newblock URL \url{https://arxiv.org/abs/2208.14023}.

\bibitem[Fragkiadaki et~al.(2015)Fragkiadaki, Levine, and Malik]{FragkiadakiLM15}
Katerina Fragkiadaki, Sergey Levine, and Jitendra Malik.
\newblock Recurrent network models for kinematic tracking.
\newblock \emph{CoRR}, abs/1508.00271, 2015.
\newblock URL \url{http://arxiv.org/abs/1508.00271}.

\bibitem[Pavllo et~al.(2018)Pavllo, Grangier, and Auli]{Pavllo2018}
Dario Pavllo, David Grangier, and Michael Auli.
\newblock Quaternet: {A} quaternion-based recurrent model for human motion.
\newblock \emph{CoRR}, abs/1805.06485, 2018.
\newblock URL \url{http://arxiv.org/abs/1805.06485}.

\bibitem[Martinez et~al.(2017)Martinez, Black, and Romero]{martinez2017humanmotionpredictionusing}
Julieta Martinez, Michael~J. Black, and Javier Romero.
\newblock On human motion prediction using recurrent neural networks, 2017.
\newblock URL \url{https://arxiv.org/abs/1705.02445}.

\bibitem[Mao et~al.(2020)Mao, Liu, Salzmann, and Li]{mao2020learningtrajectorydependencieshuman}
Wei Mao, Miaomiao Liu, Mathieu Salzmann, and Hongdong Li.
\newblock Learning trajectory dependencies for human motion prediction, 2020.
\newblock URL \url{https://arxiv.org/abs/1908.05436}.

\bibitem[Rahman et~al.(2023)Rahman, Scofano, Matteis, Flaborea, Sampieri, and Galasso]{rahman2023bestpractices2bodypose}
Muhammad Rameez~Ur Rahman, Luca Scofano, Edoardo~De Matteis, Alessandro Flaborea, Alessio Sampieri, and Fabio Galasso.
\newblock Best practices for 2-body pose forecasting, 2023.
\newblock URL \url{https://arxiv.org/abs/2304.05758}.

\bibitem[Ionescu et~al.(2014)Ionescu, Papava, Olaru, and Sminchisescu]{Human36M}
Catalin Ionescu, Dragos Papava, Vlad Olaru, and Cristian Sminchisescu.
\newblock Human3.6m: Large scale datasets and predictive methods for 3d human sensing in natural environments.
\newblock \emph{IEEE Transactions on Pattern Analysis and Machine Intelligence}, 36\penalty0 (7):\penalty0 1325--1339, 2014.
\newblock \doi{10.1109/TPAMI.2013.248}.

\bibitem[Pomerleau(1988)]{Alvinn1988}
Dean~A Pomerleau.
\newblock Alvinn: An autonomous land vehicle in a neural network.
\newblock \emph{Advances in neural information processing systems}, 1, 1988.

\bibitem[Foster et~al.(2024)Foster, Block, and Misra]{BehaviorCloningAllYouNeed2024}
Dylan~J. Foster, Adam Block, and Dipendra Misra.
\newblock Is behavior cloning all you need? understanding horizon in imitation learning, 2024.
\newblock URL \url{https://arxiv.org/abs/2407.15007}.

\bibitem[Chen et~al.(2021)Chen, Lu, Rajeswaran, Lee, Grover, Laskin, Abbeel, Srinivas, and Mordatch]{DecisionTransformer2021}
Lili Chen, Kevin Lu, Aravind Rajeswaran, Kimin Lee, Aditya Grover, Misha Laskin, Pieter Abbeel, Aravind Srinivas, and Igor Mordatch.
\newblock Decision transformer: Reinforcement learning via sequence modeling.
\newblock \emph{Advances in neural information processing systems}, 34:\penalty0 15084--15097, 2021.

\bibitem[Janner et~al.(2021)Janner, Li, and Levine]{TrajectoryTransformer2021}
Michael Janner, Qiyang Li, and Sergey Levine.
\newblock Offline reinforcement learning as one big sequence modeling problem, 2021.
\newblock URL \url{https://arxiv.org/abs/2106.02039}.

\bibitem[Wen et~al.(2022)Wen, Kuba, Lin, Zhang, Wen, Wang, and Yang]{MultiAgentRLSequence2022}
Muning Wen, Jakub~Grudzien Kuba, Runji Lin, Weinan Zhang, Ying Wen, Jun Wang, and Yaodong Yang.
\newblock Multi-agent reinforcement learning is a sequence modeling problem, 2022.
\newblock URL \url{https://arxiv.org/abs/2205.14953}.

\bibitem[Brohan et~al.(2023)Brohan, Brown, Carbajal, Chebotar, Chen, Choromanski, Ding, Driess, Dubey, Finn, Florence, Fu, Arenas, Gopalakrishnan, Han, Hausman, Herzog, Hsu, Ichter, Irpan, Joshi, Julian, Kalashnikov, Kuang, Leal, Lee, Lee, Levine, Lu, Michalewski, Mordatch, Pertsch, Rao, Reymann, Ryoo, Salazar, Sanketi, Sermanet, Singh, Singh, Soricut, Tran, Vanhoucke, Vuong, Wahid, Welker, Wohlhart, Wu, Xia, Xiao, Xu, Xu, Yu, and Zitkovich]{RT2}
Anthony Brohan, Noah Brown, Justice Carbajal, Yevgen Chebotar, Xi~Chen, Krzysztof Choromanski, Tianli Ding, Danny Driess, Avinava Dubey, Chelsea Finn, Pete Florence, Chuyuan Fu, Montse~Gonzalez Arenas, Keerthana Gopalakrishnan, Kehang Han, Karol Hausman, Alexander Herzog, Jasmine Hsu, Brian Ichter, Alex Irpan, Nikhil Joshi, Ryan Julian, Dmitry Kalashnikov, Yuheng Kuang, Isabel Leal, Lisa Lee, Tsang-Wei~Edward Lee, Sergey Levine, Yao Lu, Henryk Michalewski, Igor Mordatch, Karl Pertsch, Kanishka Rao, Krista Reymann, Michael Ryoo, Grecia Salazar, Pannag Sanketi, Pierre Sermanet, Jaspiar Singh, Anikait Singh, Radu Soricut, Huong Tran, Vincent Vanhoucke, Quan Vuong, Ayzaan Wahid, Stefan Welker, Paul Wohlhart, Jialin Wu, Fei Xia, Ted Xiao, Peng Xu, Sichun Xu, Tianhe Yu, and Brianna Zitkovich.
\newblock Rt-2: Vision-language-action models transfer web knowledge to robotic control, 2023.
\newblock URL \url{https://arxiv.org/abs/2307.15818}.

\bibitem[Kim et~al.(2024)Kim, Pertsch, Karamcheti, Xiao, Balakrishna, Nair, Rafailov, Foster, Lam, Sanketi, et~al.]{OpenVLA}
Moo~Jin Kim, Karl Pertsch, Siddharth Karamcheti, Ted Xiao, Ashwin Balakrishna, Suraj Nair, Rafael Rafailov, Ethan Foster, Grace Lam, Pannag Sanketi, et~al.
\newblock Openvla: An open-source vision-language-action model.
\newblock \emph{arXiv preprint arXiv:2406.09246}, 2024.

\bibitem[Bellemare et~al.(2013)Bellemare, Naddaf, Veness, and Bowling]{ALE}
Marc~G Bellemare, Yavar Naddaf, Joel Veness, and Michael Bowling.
\newblock The arcade learning environment: An evaluation platform for general agents.
\newblock \emph{Journal of artificial intelligence research}, 47:\penalty0 253--279, 2013.

\bibitem[Todorov et~al.(2012)Todorov, Erez, and Tassa]{mujoco}
Emanuel Todorov, Tom Erez, and Yuval Tassa.
\newblock Mujoco: A physics engine for model-based control.
\newblock In \emph{2012 IEEE/RSJ international conference on intelligent robots and systems}, pages 5026--5033. IEEE, 2012.

\bibitem[Gulino et~al.(2023)Gulino, Fu, Luo, Tucker, Bronstein, Lu, Harb, Pan, Wang, Chen, Co-Reyes, Agarwal, Roelofs, Lu, Montali, Mougin, Yang, White, Faust, McAllister, Anguelov, and Sapp]{Waymax2023}
Cole Gulino, Justin Fu, Wenjie Luo, George Tucker, Eli Bronstein, Yiren Lu, Jean Harb, Xinlei Pan, Yan Wang, Xiangyu Chen, John~D. Co-Reyes, Rishabh Agarwal, Rebecca Roelofs, Yao Lu, Nico Montali, Paul Mougin, Zoey Yang, Brandyn White, Aleksandra Faust, Rowan McAllister, Dragomir Anguelov, and Benjamin Sapp.
\newblock Waymax: An accelerated, data-driven simulator for large-scale autonomous driving research, 2023.
\newblock URL \url{https://arxiv.org/abs/2310.08710}.

\bibitem[Linderman et~al.(2017)Linderman, Johnson, Miller, Adams, Blei, and Paninski]{Linderman2017}
Scott~W. Linderman, Matthew~J. Johnson, Andrew~C. Miller, Ryan~P. Adams, David~M. Blei, and Liam Paninski.
\newblock Bayesian learning and inference in recurrent switching linear dynamical systems.
\newblock In \emph{Proceedings of the 20th International Conference on Artificial Intelligence and Statistics (AISTATS)}, volume~54 of \emph{Proceedings of Machine Learning Research}, pages 914--922, 2017.

\bibitem[Batty et~al.(2019)Batty, Whiteway, Saxena, Biderman, Abe, Musall, Gillis, Markowitz, Churchland, Cunningham, Datta, Linderman, and Paninski]{Batty2019}
Eleanor Batty, Matthew Whiteway, Shreya Saxena, Dan Biderman, Taiga Abe, Simon Musall, Winthrop Gillis, Jeffrey Markowitz, Anne Churchland, John~P. Cunningham, Sandeep~R. Datta, Scott Linderman, and Liam Paninski.
\newblock {BehaveNet}: nonlinear embedding and {Bayesian} neural decoding of behavioral videos.
\newblock In \emph{Advances in Neural Information Processing Systems}, volume~32, pages 15680--15691, 2019.

\bibitem[Weinreb et~al.(2024)Weinreb, Pearl, Lin, Osman, Zhang, Annapragada, Conlin, Hoffmann, Makowska, Gillis, Jay, Ye, Mathis, Mathis, Pereira, Linderman, and Datta]{Weinreb2024}
Caleb Weinreb, Jonah~E. Pearl, Sherry Lin, Mohammed Abdal~Monium Osman, Libby Zhang, Sidharth Annapragada, Eli Conlin, Red Hoffmann, Sofia Makowska, Winthrop~F. Gillis, Maya Jay, Shaokai Ye, Alexander Mathis, Mackenzie~W. Mathis, Talmo Pereira, Scott~W. Linderman, and Sandeep~Robert Datta.
\newblock Keypoint-{MoSeq}: parsing behavior by linking point tracking to pose dynamics.
\newblock \emph{Nature Methods}, 21\penalty0 (7):\penalty0 1329--1339, 2024.
\newblock \doi{10.1038/s41592-024-02318-2}.

\bibitem[Calhoun et~al.(2019)Calhoun, Pillow, and Murthy]{Calhoun2019}
Adam~J. Calhoun, Jonathan~W. Pillow, and Mala Murthy.
\newblock Unsupervised identification of the internal states that shape natural behavior.
\newblock \emph{Nature Neuroscience}, 22\penalty0 (12):\penalty0 2040--2049, 2019.
\newblock \doi{10.1038/s41593-019-0533-x}.

\bibitem[Vaxenburg et~al.(2025)Vaxenburg, Siwanowicz, Merel, Robie, Morrow, Novati, Stefanidi, Both, Card, Reiser, et~al.]{flybody2025}
Roman Vaxenburg, Igor Siwanowicz, Josh Merel, Alice~A Robie, Carmen Morrow, Guido Novati, Zinovia Stefanidi, Gert-Jan Both, Gwyneth~M Card, Michael~B Reiser, et~al.
\newblock Whole-body physics simulation of fruit fly locomotion.
\newblock \emph{Nature}, 643\penalty0 (8074):\penalty0 1312--1320, 2025.

\bibitem[Aldarondo et~al.(2024)Aldarondo, Merel, Marshall, Hasenclever, Klibaite, Gellis, Tassa, Wayne, Botvinick, and {\"O}lveczky]{virtualrodent2024}
Diego Aldarondo, Josh Merel, Jesse~D Marshall, Leonard Hasenclever, Ugne Klibaite, Amanda Gellis, Yuval Tassa, Greg Wayne, Matthew Botvinick, and Bence~P {\"O}lveczky.
\newblock A virtual rodent predicts the structure of neural activity across behaviours.
\newblock \emph{Nature}, 632\penalty0 (8025):\penalty0 594--602, 2024.

\bibitem[Lappalainen et~al.(2024)Lappalainen, Tschopp, Prakhya, McGill, Nern, Shinomiya, Takemura, Gruntman, Macke, and Turaga]{Lappalainen2024}
Janne~K Lappalainen, Fabian~D Tschopp, Sridhama Prakhya, Mason McGill, Aljoscha Nern, Kazunori Shinomiya, Shin-ya Takemura, Eyal Gruntman, Jakob~H Macke, and Srinivas~C Turaga.
\newblock Connectome-constrained networks predict neural activity across the fly visual system.
\newblock \emph{Nature}, 634\penalty0 (8036):\penalty0 1132--1140, 2024.

\bibitem[Eyjolfsdottir et~al.(2016)Eyjolfsdottir, Branson, Yue, and Perona]{Eyrun2016}
Eyrun Eyjolfsdottir, Kristin Branson, Yisong Yue, and Pietro Perona.
\newblock Learning recurrent representations for hierarchical behavior modeling, 2016.
\newblock URL \url{https://arxiv.org/abs/1611.00094}.

\bibitem[Im et~al.(2020)Im, Kwak, and Branson]{ImBranson}
Daniel~Jiwoong Im, Iljung Kwak, and Kristin Branson.
\newblock Evaluation metrics for behaviour modeling.
\newblock \emph{arXiv preprint arXiv:2007.12298}, 2020.

\bibitem[Robie et~al.(2024)Robie, Taylor, Schretter, Kabra, and Branson]{flydisco}
Alice~A Robie, Adam~L Taylor, Catherine~E Schretter, Mayank Kabra, and Kristin Branson.
\newblock The fly disco: hardware and software for optogenetics and fine-grained fly behavior analysis.
\newblock \emph{bioRxiv}, pages 2024--11, 2024.

\bibitem[Branson et~al.(2026)Branson, Robie, and Schretter]{FlyMABe2022v3}
Kristin Branson, Alice~A. Robie, and Catherine Schretter.
\newblock {FlyMABe2022} (revision c705992), 2026.
\newblock URL \url{https://huggingface.co/datasets/kristinbranson/FlyMABe2022}.

\bibitem[Marr(1982)]{Marr1982}
David Marr.
\newblock \emph{Vision: A Computational Investigation into the Human Representation and Processing of Visual Information}.
\newblock W. H. Freeman, San Francisco, 1982.

\bibitem[Gruntman et~al.(2018)Gruntman, Romani, and Reiser]{Gruntman2018}
Eyal Gruntman, Sandro Romani, and Michael~B. Reiser.
\newblock Simple integration of fast excitation and offset, delayed inhibition computes directional selectivity in {Drosophila}.
\newblock \emph{Nature Neuroscience}, 21\penalty0 (2):\penalty0 250--257, 2018.
\newblock \doi{10.1038/s41593-017-0046-4}.

\bibitem[Ito et~al.(2014)Ito, Shinomiya, Ito, Armstrong, Boyan, Hartenstein, Harzsch, Heisenberg, Homberg, Jenett, et~al.]{ito2014systematic}
Kei Ito, Kazunori Shinomiya, Masayoshi Ito, J~Douglas Armstrong, George Boyan, Volker Hartenstein, Steffen Harzsch, Martin Heisenberg, Uwe Homberg, Arnim Jenett, et~al.
\newblock A systematic nomenclature for the insect brain.
\newblock \emph{Neuron}, 81\penalty0 (4):\penalty0 755--765, 2014.

\bibitem[Currier and Nagel(2020)]{CurrierNagel2020}
Timothy~A. Currier and Katherine~I. Nagel.
\newblock Multisensory control of navigation in the fruit fly.
\newblock \emph{Current Opinion in Neurobiology}, 64:\penalty0 10--16, 2020.
\newblock \doi{10.1016/j.conb.2019.11.017}.

\bibitem[Okray et~al.(2023)Okray, Jacob, Stern, Desmond, Otto, Talbot, Vargas-Gutierrez, and Waddell]{Okray2023}
Zeynep Okray, Pedro~F. Jacob, Ciara Stern, Kieran Desmond, Nils Otto, Clifford~B. Talbot, Paola Vargas-Gutierrez, and Scott Waddell.
\newblock Multisensory learning binds neurons into a cross-modal memory engram.
\newblock \emph{Nature}, 617:\penalty0 777--784, 2023.
\newblock \doi{10.1038/s41586-023-06013-8}.

\bibitem[George et~al.(2024)George, Rastogi, de~Cothi, Clopath, Stachenfeld, and Barry]{RatInABox}
Tom~M George, Mehul Rastogi, William de~Cothi, Claudia Clopath, Kimberly Stachenfeld, and Caswell Barry.
\newblock Ratinabox, a toolkit for modelling locomotion and neuronal activity in continuous environments.
\newblock \emph{Elife}, 13:\penalty0 e85274, 2024.

\bibitem[Wang et~al.(2021)Wang, Xu, Narasimhan, and Wang]{NEURIPS2021_2fd5d41e}
Jiashun Wang, Huazhe Xu, Medhini Narasimhan, and Xiaolong Wang.
\newblock Multi-person 3d motion prediction with multi-range transformers.
\newblock In M.~Ranzato, A.~Beygelzimer, Y.~Dauphin, P.S. Liang, and J.~Wortman Vaughan, editors, \emph{Advances in Neural Information Processing Systems}, volume~34, pages 6036--6049. Curran Associates, Inc., 2021.
\newblock URL \url{https://proceedings.neurips.cc/paper_files/paper/2021/file/2fd5d41ec6cfab47e32164d5624269b1-Paper.pdf}.

\bibitem[Wilson(1966)]{wilson1966insect}
Donald~M Wilson.
\newblock Insect walking.
\newblock \emph{Annual review of entomology}, 11\penalty0 (1):\penalty0 103--122, 1966.

\bibitem[Kabra et~al.(2013)Kabra, Robie, Rivera-Alba, Branson, and Branson]{kabra2013jaaba}
Mayank Kabra, Alice~A Robie, Marta Rivera-Alba, Steven Branson, and Kristin Branson.
\newblock Jaaba: interactive machine learning for automatic annotation of animal behavior.
\newblock \emph{Nature methods}, 10\penalty0 (1):\penalty0 64--67, 2013.

\bibitem[Zhang et~al.(2023)Zhang, Zhang, Cun, Huang, Zhang, Zhao, Lu, and Shen]{zhang2023t2mgptgeneratinghumanmotion}
Jianrong Zhang, Yangsong Zhang, Xiaodong Cun, Shaoli Huang, Yong Zhang, Hongwei Zhao, Hongtao Lu, and Xi~Shen.
\newblock T2m-gpt: Generating human motion from textual descriptions with discrete representations, 2023.
\newblock URL \url{https://arxiv.org/abs/2301.06052}.

\bibitem[Tevet et~al.(2022)Tevet, Raab, Gordon, Shafir, Cohen-Or, and Bermano]{tevet2022humanmotiondiffusionmodel}
Guy Tevet, Sigal Raab, Brian Gordon, Yonatan Shafir, Daniel Cohen-Or, and Amit~H. Bermano.
\newblock Human motion diffusion model, 2022.
\newblock URL \url{https://arxiv.org/abs/2209.14916}.

\bibitem[Eyjolfsdottir et~al.(2014)Eyjolfsdottir, Branson, Burgos-Artizzu, Hoopfer, Schor, Anderson, and Perona]{FlyTracker}
Eyrun Eyjolfsdottir, Steve Branson, Xavier~P. Burgos-Artizzu, Eric~D. Hoopfer, Jonathan Schor, David~J. Anderson, and Pietro Perona.
\newblock Detecting social actions of fruit flies.
\newblock In \emph{European Conference on Computer Vision (ECCV)}, volume 8690 of \emph{Lecture Notes in Computer Science}, pages 772--787. Springer, 2014.
\newblock \doi{10.1007/978-3-319-10605-2_50}.

\end{thebibliography}
}


\appendix

\section{Supplementary videos}

Supplementary videos are available on Figshare:\\
\url{\fsdoi}

\href{\vidRef}{\textbf{Video 1}} Example simulations using \textsc{Reference} model. Simulated agents are highlighted and in bold. In this video only male flies are simulated, the ohters are female.

\href{\vidRealData}{\textbf{Video 2}} Real fly trajectories corresponding to the ones in Video 1. Notice that the females in Video 1 have the same trajectories as in Video 2 and the males start in the same location at frame 0 and then the simulated agents diverge from the real ones.

\textbf{Video 3 - 8} Same as Video 1 but with simulations using the following models: \href{\vidShortCtx}{\textsc{Shorter context} (3)}, \href{\vidNoDisc}{\textsc{No discretization} (4)}, \href{\vidDiscAll}{\textsc{Discretize all} (5)}, \href{\vidStaticPose}{\textsc{Static local pose} (6)}, \href{\vidNoHandPose}{\textsc{No handcrafted pose} (7)},  \href{\vidKeypoints}{\textsc{Keypoints} (8)}.

\href{\vidFullRef}{\textbf{Video 9}} Example simulations from \textsc{Reference} model on full MABe 2022 dataset. Here all agents are simulated. Females are colored in matplotlib's 'spring' color palette (pinks and yellows) and males in the 'winter' palette (blues and greens). Males are also plotted with darker skeleton edges. 

\href{\vidFullRealData}{\textbf{Video 10}} Real fly trajectories corresponding to those in video 9. The first position of all flies is the same between the two videos and then the simulated trajectories diverge. 

\href{\vidRatInABox}{\textbf{Video 11}} Example trajectories from the RatInABox experiment described in \ref{sec:appendix_ratinabox}. This is a video representation of \cref{fig:ratinabox_results}b.

\section{Experiment details and additional results}
\label{sec:appendix_additional_results}

\subsection{Qualitative evaluation}
\label{sec:appendix_qualitative_evaluation}

\begin{figure}[htb]
    \centering
    \makebox[\textwidth][c]{%
        \includegraphics[width=1.0\linewidth]{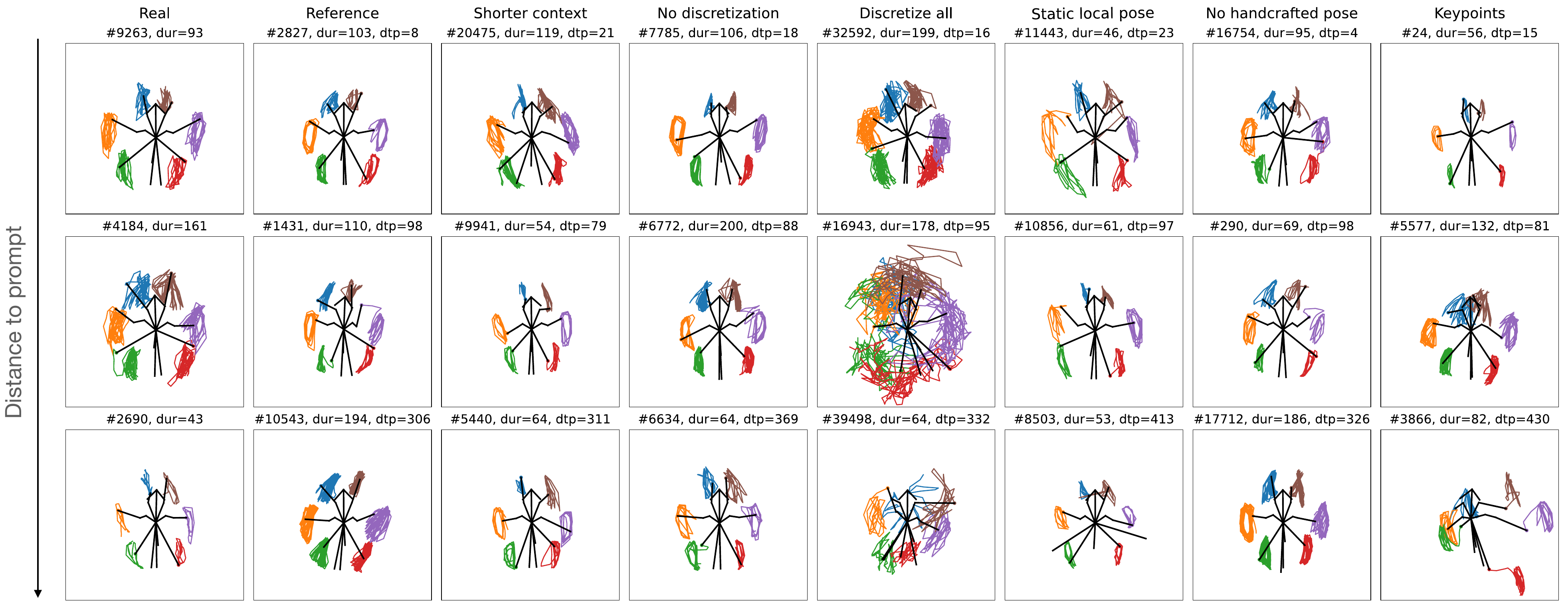}
    }
    \vspace{-4mm}
    \caption{Relative leg-tip positions for example walking bouts for ground truth and for simulated data from all model variants. All variants have learned the tripod gait pattern and can produce it deep into the rollout, though \discretizeAllModel and \keypointsModel increasingly deviate from plausible walking as the simulation progresses.}
    \label{fig:walk_bouts_all}
\end{figure}

\cref{fig:walk_bouts_all} shows trajectories of leg tips for randomly selected example walking bouts, in the flies frame of reference. The bouts are listed in increasing distance to the prompt (DTP), from top to bottom. The first column shows example bouts from the real data, the remaining columns are from trajectories that were simulated with different model variants. For most of the models we see that walking bouts look similar to the real ones, even at distances as far as 413 frames from the prompt. This indicates that the models have learned a walking pattern which is executed well into the simulation. The models that we see fail this qualitative test are \textsc{Discretize all} and \textsc{Keypoints}.

\subsection{Feature distribution comparison}
\label{sec:appendix_feature_distribution_comparison}

\begin{figure}[htb]
    \centering
    \makebox[\textwidth][c]{%
        \includegraphics[width=0.95\linewidth]{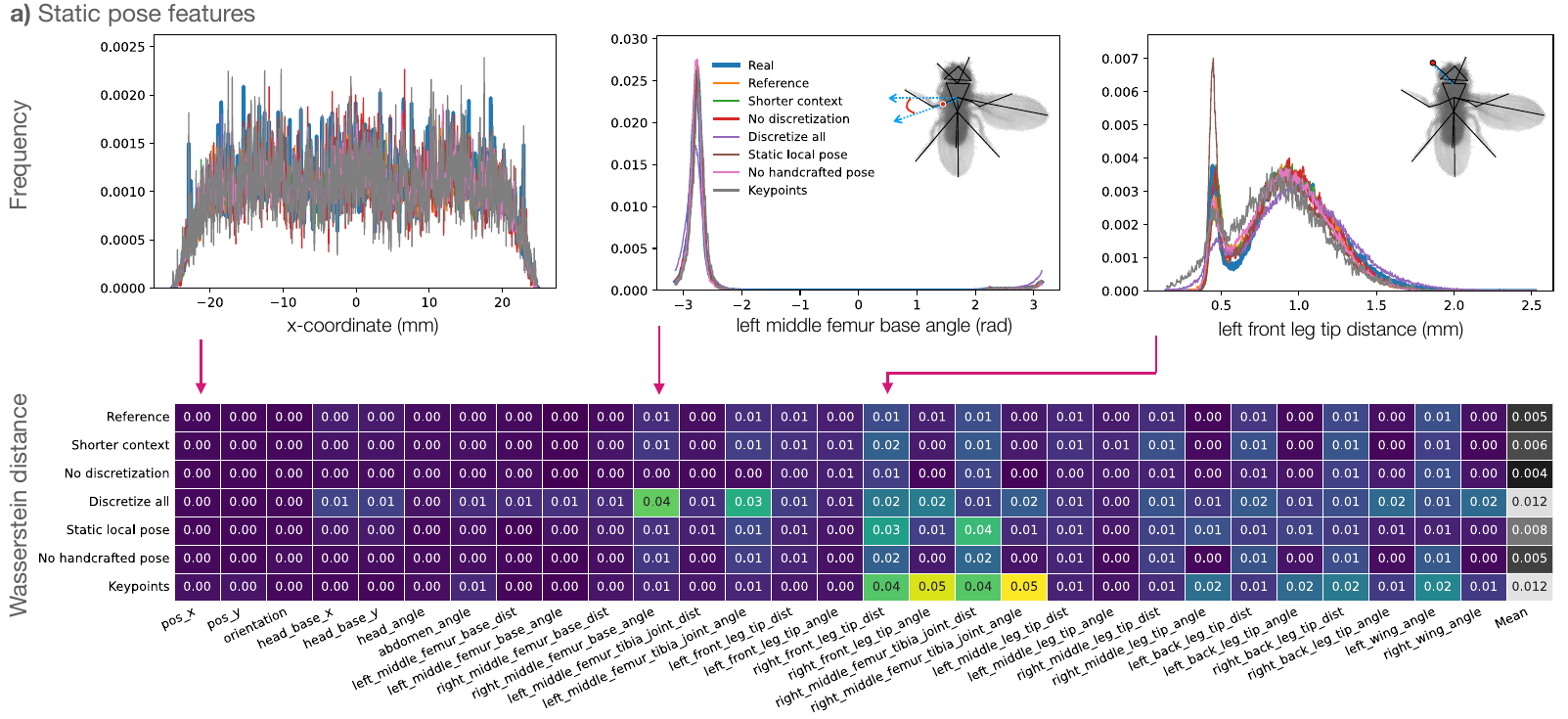}        
    }
    \makebox[\textwidth][c]{%
        \includegraphics[width=0.95\linewidth]{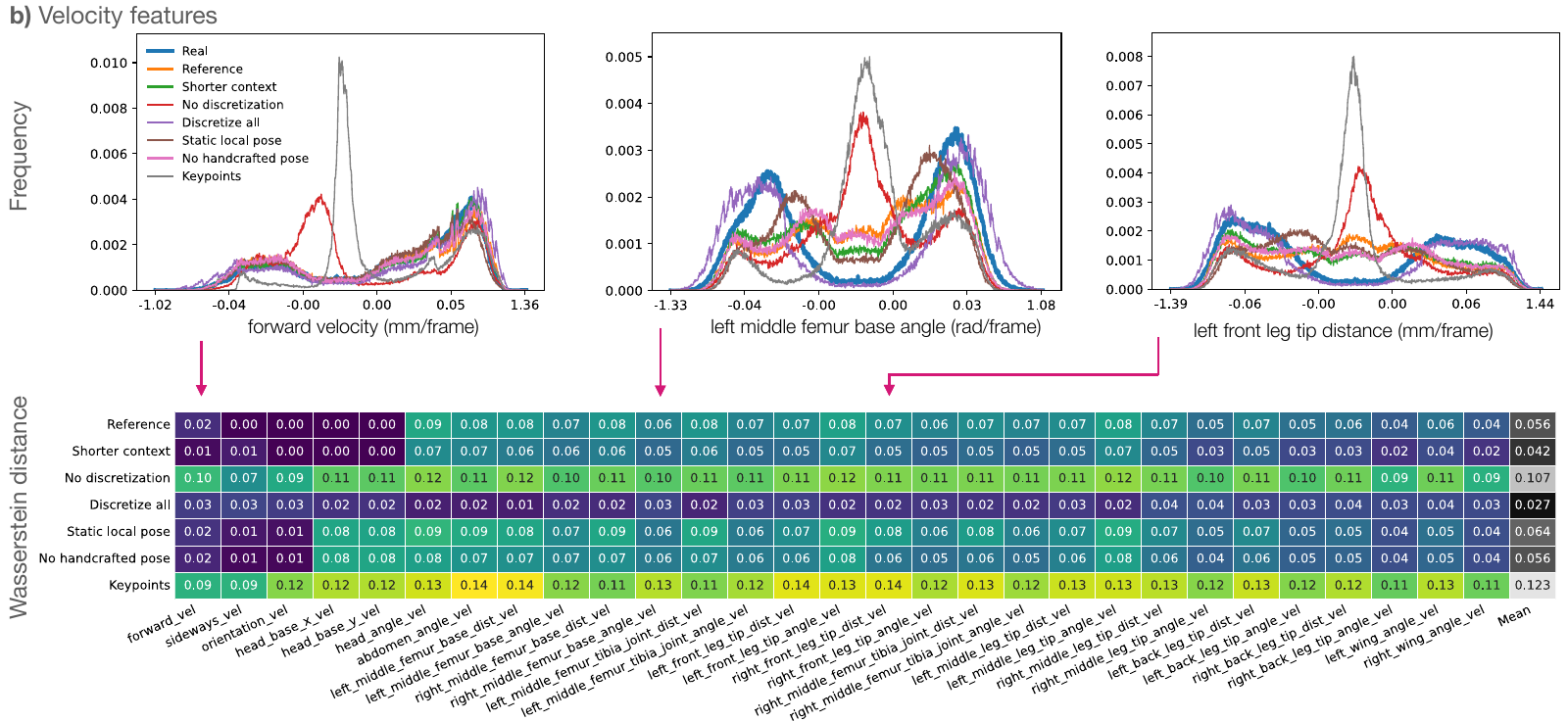}        
    }
    \vspace{-4mm}
    \caption{\textbf{a}. distribution of three selected pose velocity features, real vs. simulated from each model variant, with Wasserstein distance reported across all position/pose features. \textbf{b}. Same, for position/pose features, instead of velocity. Thin vertical lines indicate the linear portion of the symlog transform.}
    \label{fig:wasser_full}
 \end{figure}

\label{sec:wasserstein_details}

We compare the marginal distribution of each of $58$ interpretable features ---
the $29$ static pose features (global position $x$, $y$, orientation, and the $26$
local-pose features) and their $29$ velocities --- between real and simulated data. For velocity features, whose magnitudes span several orders of magnitude, we first apply a symlog transform with linear threshold equal to $1\%$ of that feature's standard deviation in the real data. For each feature we build a $1{,}000$-bin histogram for both the real and simulated data, with bin edges linearly spaced between the minimum and maximum of the real data. We then compute the Wasserstein distance between the two histograms, normalized by the histogram's support. 

\cref{fig:wasser_full} shows the Wasserstein distance for all pose features and velocity features, comparing the real data with data simulated for up to 64 frames from the prompt. Static pose distributions (\cref{fig:wasser_full}a) are close to real across nearly all features and model variants, with a few exceptions. Distribution distances for the front-leg-tip-related feature distributions for the \keypointsModel model variants are higher, and, to a lesser degree, several local pose features for the \discretizeAllModel and \staticLocalPoseModel variants. Some of these errors are due to distribution drift during rollout: errors result in poses outside those in the training data domain, and the models generalize poorly, resulting in even bigger errors. For example, for the left-front-leg-tip distance, we see values close to 0 mm that are never seen in real data.

Figure~\ref{fig:feat_per_dtp} (bottom) shows how mean forward velocity, distance to arena center, and distance to the nearest fly as a function of distance to the prompt. Ideally, the mean would remain near the real distribution mean throughout. \referenceModel and \noHandcraftedPoseModel are similar and remain closest to the real distribution deep into the rollout. The main failure is a slight decrease in the mean forward velocity toward the end of the rollout. In contrast, we see failure modes of other model variants: \discretizeAllModel distances increase over distance from prompt, which corresponds to the erratic behavior we see as the rollout progresses; \keypointsModel, \noDiscretizationModel, and \staticLocalPoseModel all have mean forward velocity going to zero as rollout progresses, corresponding to the observation that these simulated flies don't move. While by eye we didn't notice a strong difference between \referenceModel and \shorterContextModel, we see a clear difference in the mean distance to arena center and distance to closest fly in the \shorterContextModel, indicating that the longer context is helpful in reproducing this behavior feature. 

\subsection{Real-vs-simulated discriminator}
\label{sec:appendix_real_vs_sim}
\label{sec:appendix_discriminator_details}

We train discriminators on the 29-dimensional, instantaneous static pose, and on the instantaneous pose velocity. A discriminator accuracy of 0.5 indicates the simulated data is indistinguishable from real data; an accuracy of 1.0 indicates perfect separability. The discriminator is a multilayer perceptron with hidden layers of width $128$, $256$, and $128$ (ReLU activations and a single sigmoid output), trained with binary cross-entropy (Adam, learning rate $10^{-3}$, $3$ epochs) to classify each frame's feature vector as real ($0$) or simulated ($1$); inputs are standardized to zero
mean and unit variance. Real and simulated frames are pooled and split $80/20$ into train/test, and accuracy is the fraction of held-out frames classified correctly at a threshold of $0.5$ (so $0.5$ is indistinguishable and $1.0$ perfectly separable). As a control, a discriminator trained to separate two halves of the real data alone achieves $\approx 0.5$. For the per-feature and feature-group analyses the same discriminator is trained on individual features or feature subsets; for the distance-to-prompt analysis the training frames are restricted to a given DTP range.

\begin{figure}[htb]
    \centering
    \includegraphics[width=.95\linewidth]{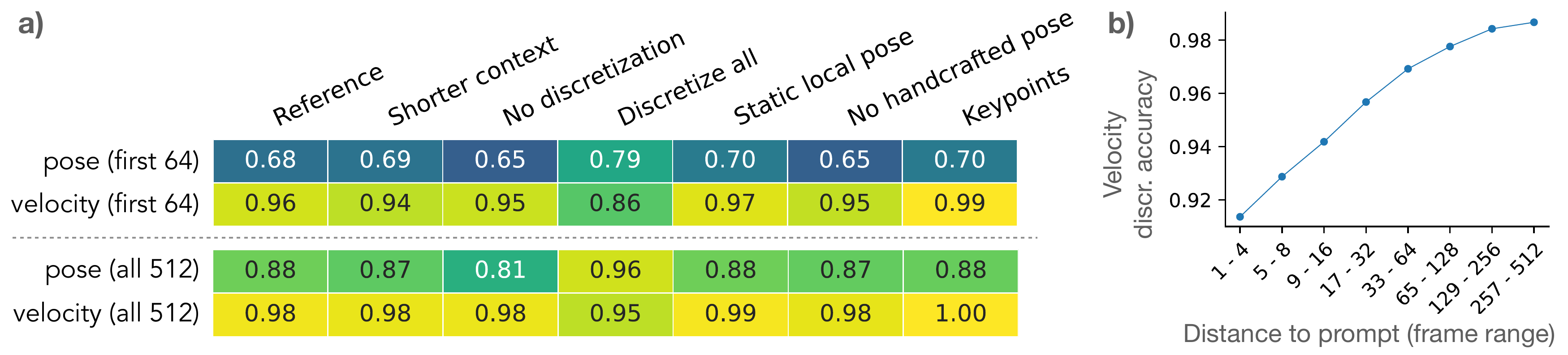}
    \caption{\textbf{a}. Real-vs-simulated discrimination accuracy using position and pose (top) and velocity
    (bottom) features over the first 64 (top) and 512 (bottom) simulated frames. \textbf{b}. Discrimination accuracy for
    data simulated with the \referenceModel model, using velocity features at increasing
    distance from the prompt.}
    \label{fig:discr_all}
\end{figure}

\cref{fig:discr_all} shows discriminator accuracy trained on the static position and pose (top) and instantaneous velocity (bottom) features, using the first 64 (top) and all 512 (bottom) simulated frames. For the \referenceModel model, over the first 64 frames, the pose-based discriminator achieves an accuracy as low as 0.65, whereas the velocity-based discriminator is substantially more accurate. One explanation is that a stationary fly copies the original pose -- itself a valid real pose -- even though it is not really moving like a real fly. This explains the good performance of the \noDiscretizationModel variant here, which suffers from a strong bias toward predicting no movement, sometimes even for the entire rollout. Perhaps surprisingly, the \keypointsModel model, which also primarily outputs no motion, performs slightly worse than our \referenceModel model, while the \noHandcraftedPoseModel model performs better. All model variants perform worse on fooling the velocity discriminator than the pose discriminator, with \discretizeAllModel surprisingly performing best, despite sampling from all features independently during rollout. The strong performance of the velocity-based discriminator suggests subtle discrepancies between real and simulated movement that are not obvious to the expert eye. As a sanity check, discriminators trained on real-vs-real data achieve accuracy close to 0.50. Training a discriminator on simulations from the \referenceModel model at varying distances from the prompt (Figure~\ref{fig:discr_all}b), we find that even the first four simulated frames yield an accuracy of 0.92.

\begin{figure}[htb]
\centering
\includegraphics[width=.8\linewidth]{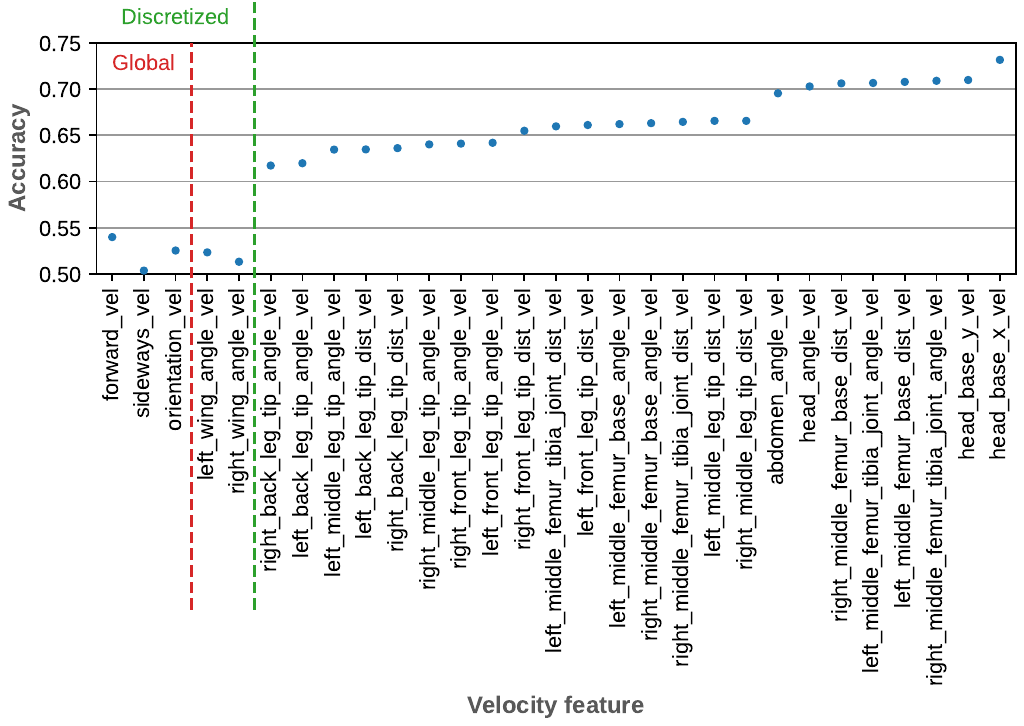}
\caption{Real-vs-simulated discrimination accuracy using individual per-frame
    pose-velocity features. Extension of data in \cref{fig:discr_per_feat}c.}
\label{fig:discr_per_feat_all}
\end{figure}

\cref{fig:discr_per_feat_all} shows the accuracy of discriminators trained on a single velocity feature. To understand the velocity discriminator accuracy better, we train discriminators using individual pose-velocity features as well as feature groups (e.g. all global features, both wing angles, all continuous features) for the REFERENCE model. The discriminator performs close to chance on discretized features individually (0.52–0.53), consistent with how closely those distributions match (\ref{fig:wasser_full}). However, jointly considering the global features raises accuracy to 0.67 (\ref{fig:discr_per_feat}), indicating that these features are not well coordinated with one another during simulation. The global velocity features alone reach only 0.58, a modest increase over the individual features, while the two wing angles together reach 0.63 – a much larger jump, indicating that wing angle coordination is particularly poorly captured. As a control, we compared the two continuous features with the lowest individual error (back-leg tip angle velocities): considered independently they reach 0.59, and jointly 0.63, a far smaller gap than for the wing angles.

\subsection{Behavior-pattern frequency}
\label{sec:appendix_frequency_details}

The courtship behaviors (\textbf{courting}, \textbf{chasing}, \textbf{wing extension}) are classified using JAABA \citep{kabra2013jaaba} applied to trajectory tracking data. Classifiers were trained on the manually annotated data from the MABe 2022 fly dataset, and input both keypoint trajectories and ellipses fit to the fly body using \cite{FlyTracker}. Balanced accuracy on the test data are in \cref{table:jaaba_accuracy}. For simulated data, we only have keypoints available, thus approximated the ellipses with a regression fit from the keypoints:
\begin{align*}
  \mathbf{c} &= \left(\mathbf{p}_{\text{head}} + \mathbf{p}_{\text{abdomen}}\right) / 2 \\
  \theta &= \operatorname{atan2}\!\left(y_{\text{head}} - y_{\text{abdomen}},\; x_{\text{head}} - x_{\text{abdomen}}\right) \\
  a &= 0.4772\,\left\lVert \mathbf{p}_{\text{head}} - \mathbf{p}_{\text{abdomen}} \right\rVert + 0.0556 \\
  b &= 0.6890\,\left\lVert \mathbf{p}_{\text{thorax},L} - \mathbf{p}_{\text{thorax},R} \right\rVert + 0.000
\end{align*}
The keypoint-based ellipse fit slightly decreased validation accuracy (\cref{table:jaaba_accuracy}).

\begin{table}[htb]
  \centering
  \label{tab:ellipse_accuracy}
  \begin{tabular}{lcc}
    \toprule
    & \multicolumn{2}{c}{Balanced accuracy} \\
    \cmidrule(lr){2-3}
    Behavior & FlyTracker ellipses & Approx.\ keypoint-based ellipses \\
    \midrule
    Chase     & 0.863 & 0.823 \\
    Wing ext. & 0.915 & 0.837 \\
    Courtship & 0.962 & 0.831 \\
    \bottomrule
  \end{tabular}
  \caption{Balanced accuracy of JAABA-trained behavior classifiers with FlyTracker-based ellipses and keypoint-based ellipses approximations. }
  \label{table:jaaba_accuracy}
\end{table}

\textbf{Walking} is detected by a handcrafted rule on the centroid (thorax-keypoint)
speed: a bout is a run of frames whose $\log$-speed exceeds $0$ (i.e.\ speed
$\gtrsim 1$~px/frame) sustained for more than $20$ consecutive frames. 

For each behavior we report the relative frequency error $RE = (f_\text{real} - f_\text{sim}) / f_\text{real}$ between the real and simulated occurrence frequencies. 

\label{sec:appendix_evolution_of_behavior_in_simulation}

\begin{figure}[htb]
    \centering
    \includegraphics[width=1.0\linewidth]{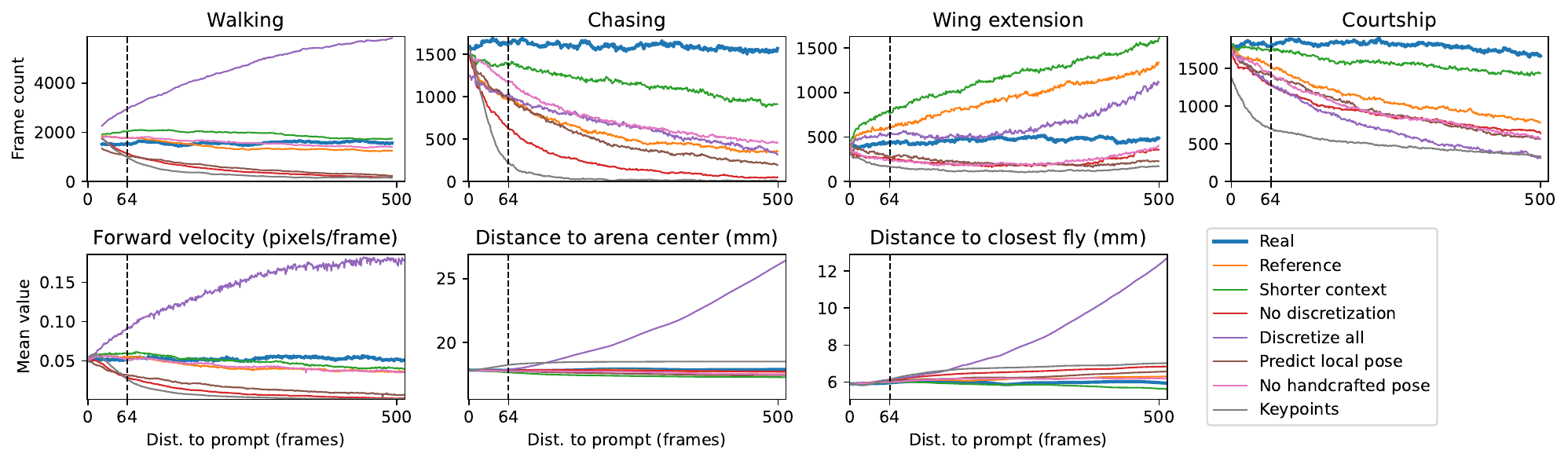}
    \vspace{-4mm}
    \caption{Evolution of behavior frequency (top) and selected feature means (bottom) over simulation duration. }
    \label{fig:feat_per_dtp}
\end{figure}

In our evaluation we compare of feature distribution and behavior frequency using the first 64 frames of simulation. This threshold was chosen to reduce the number of frames where simulation has gone catastrophically bad. Empirically we found that the first 64 frames looked qualitatively convincing. To get a sense for how things develop over the course of simulation we plot in \cref{fig:feat_per_dtp} the evolution of behavior class frequencies (top) and per-frame feature means (bottom) as function of distance to prompt. The thick blue line shows the ground truth, which is roughly fixed over this duration, as we would expect as this is averaged over thousands of trajectories, and for real trajectories, DTP has no meaning. \textsc{Discretize all} notably deviaties the most from the real data. \textsc{Keypoints} and \textsc{No discretization} are the lowest in terms of chasing, walking and velocity, which is not surprising as both of those models do not discretize the features and flies become mostly stationary. 

\subsection{Linear probes of model internals}
\label{sec:appendix_probes}

To ask whether the model's internal representation encodes higher-level labels, we
take the $2048$-dimensional hidden state at the current frame from the $6$th of the
$10$ transformer layers and train a linear classifier (logistic regression: a single
linear layer with sigmoid output, binary cross-entropy, Adam learning rate $10^{-4}$) to predict each binary label. Labels include fly-type/genotype categories and expert-annotated per-frame behavior categories. Classifiers are trained on the training set and evaluated on the validation set, and scored with the Matthews correlation coefficient (main text) because the labels are highly imbalanced. All fly data is restricted to courting male flies.

Because labels are highly imbalanced, we evaluate classifiers using the Matthews correlation coefficient (MCC):
\begin{equation}
   \text{MCC} = \frac{TP \cdot TN - FP \cdot FN}{\sqrt{(TP+FP)(TP+FN)(TN+FP)(TN+FN)}},
\end{equation}
where TP, FP, TN, and FN are true positives, false positives, true negatives, and false
negatives, respectively. An MCC of 0 corresponds to chance performance, $+1$ to perfect
prediction, and $-1$ to total disagreement. Relative to other common metrics, MCC is more
robust to class imbalance than F1 (which ignores the true-negative rate) or balanced accuracy (which underweights precision). We note that these numbers are not directly comparable to those in the MABe 2022 challenge in part because we are working with a subset of the data. We train and evaluate only on courting male flies, and e.g.~differentiating 71G01 flies from other courting lines is a much more difficult task than differentiating courting from non-courting flies. 

\subsection{MABe 2022 challenge} 

\begin{figure}[htb]
    \centering
    \includegraphics[width=0.8\linewidth]{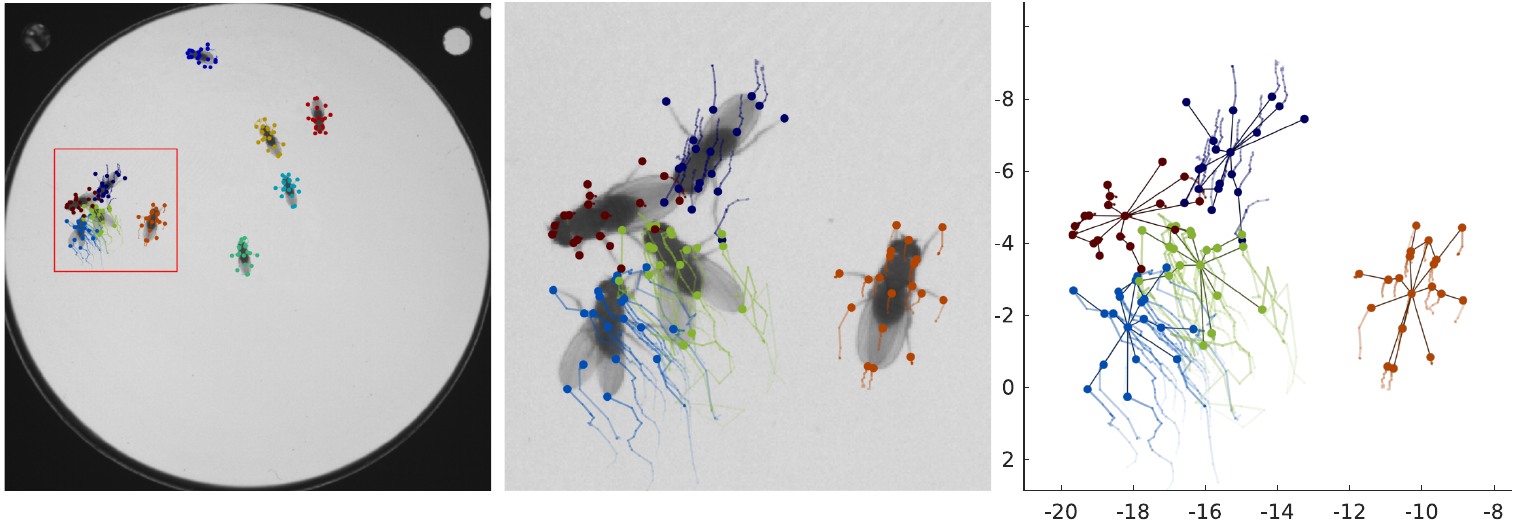}
    \caption{MABe 2022 fly behavior data video frame and trajectories.}
    \label{fig:mabe_fly_data}
\end{figure}

\begin{figure}
    \centering
    \includegraphics[width=1.0\linewidth]{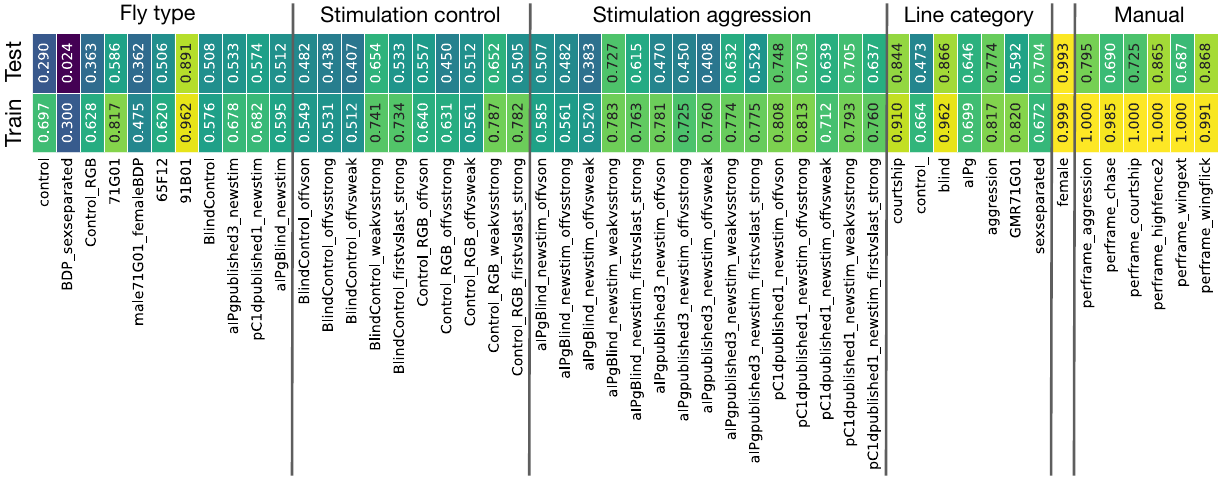}
    \caption{F1 score for training and test data, for classifiers trained on all labels from the MABe 2022 dataset.}
    \label{fig:mabe_f1_all}
\end{figure}

In our results section \ref{sec:results} we described the MABe 2022 challenge, which involves producing a data embedding and training low dimensional classifiers for labels, never seen by the embedding model, to test whether information useful for classifying behavior or fly types is captured by the embedding. In \cref{fig:mabe_comparison} we showed our results summarized by broader categories of labels. Here, in \cref{fig:mabe_f1_all} we show the full breakdown of the F1 score per behavior, both on training and test data. The results on the training data give an indication of how separable the data is by each label. 

\subsection{Prompt test}
\label{sec:appendix_prompt_test}

\begin{figure}
    \centering
    \includegraphics[width=1.0\linewidth]{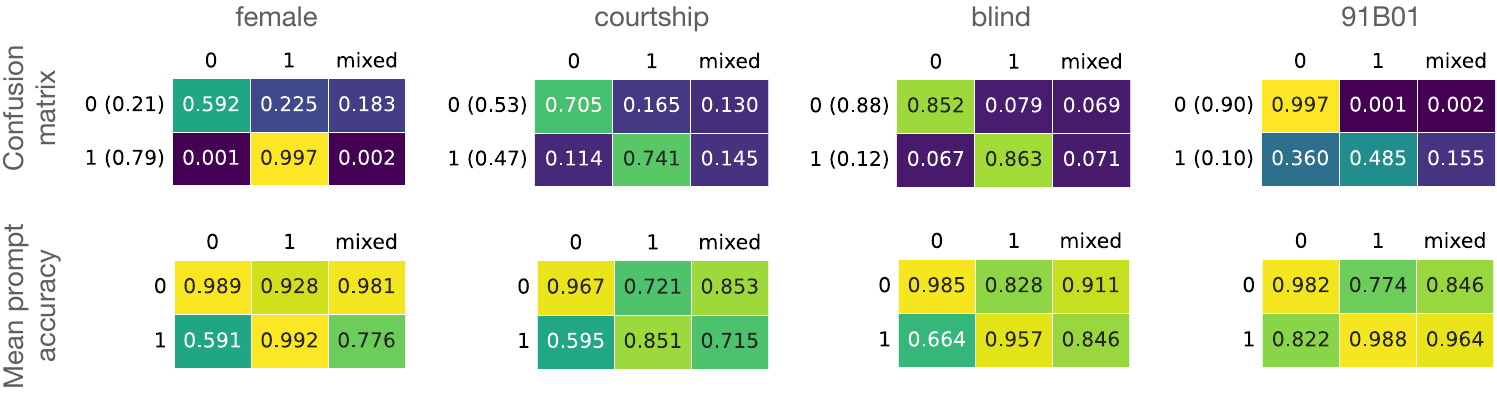}
    \caption{\textbf{Top.} Confusion matrices showing ground truth class of a prompt (rows) and the classification of the corresponding simulated track (cols). The fraction of gt labels is represented in brackets. \textbf{Bottom.} Mean accuracy of classifier on prompts.}
    \label{fig:prompt_confusion}
\end{figure}

\begin{figure}
    \centering
    \includegraphics[width=1.0\linewidth]{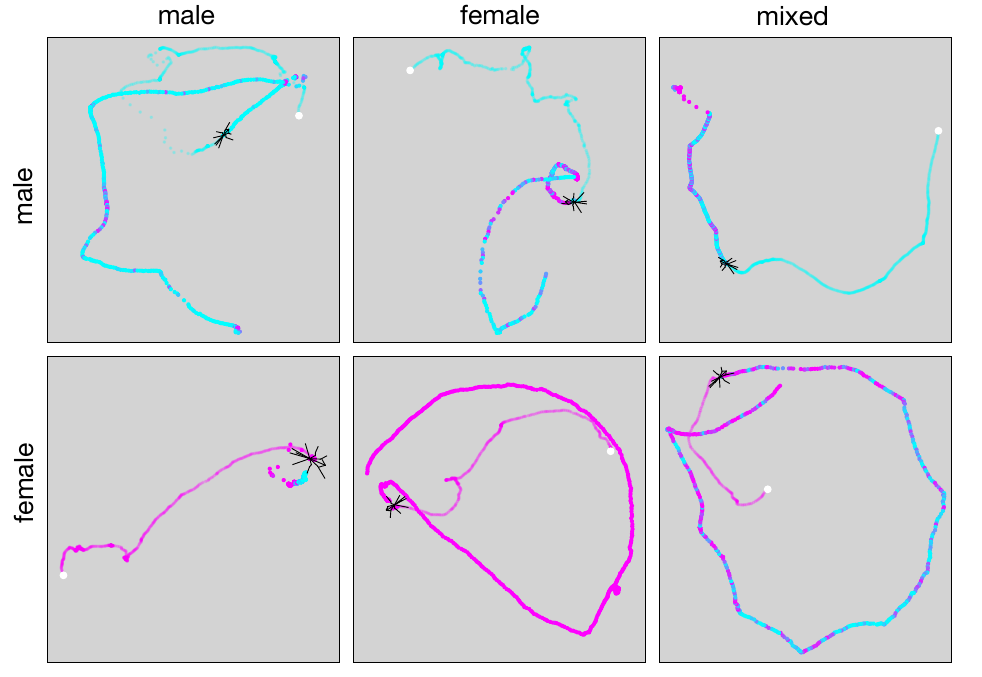}
    \caption{Example trajectories corresponding to the confusion matrix. Trajectory color indicates male (cyan) - female (magenta) score, white dot the start of a prompt, and fly skeleton the start of a simulation.}
    \label{fig:prompt_mf_tracks}
\end{figure}

We are also very interested in knowing: when simulation is prompted with a female trajectory, does it generate a simulated trajectory that is also classified as female? To test this we compare the label of prompts to the classification of the simulation track that follows it. Classification is done on a per-frame basis, so to determine the label for the simulated trajectory we consider the majority vote. If the fraction of votes is between 0.4 and 0.6 we classify it as mixed. \cref{fig:prompt_confusion} shows the confusion matrix for four categories: The fly-type \emph{91B01}, the line categories \emph{courtship} and \emph{blind}, and the \emph{female} category. We chose these categories because they have higher than 80\% accuracy on the test data. All categories contain 32822 sequences except \emph{female} which has 52790. Below the confusion matrices are corresponding matrices showing the mean accuracy of the classifier on the prompts. This shows generated sequences with label different from their prompts tend to have lower classification accuracy on the prompts, indicating that these sequences might not represent well that class. In \cref{fig:prompt_mf_tracks} we show trajectories for one random sample from each entry in the female confusion matrix. 

\section{Library design}
\label{sec:appendix_library_design}

Our Python/PyTorch-based library is built around the observation that, to train and generate from agent-centric models, the \emph{same data} needs to be available in \emph{many representations}: world-frame, egocentric, sensory, ML-friendly, and translating between these representations requires composing sequences of operations or their inverses. Our library makes these operations explicit, manipulable objects that are composable, invertible, and serializable alongside the model. The sequence of operations applied to a chunk of data is itself stored, so the library knows how a given representation was constructed and can invert or serialize it on demand. 

\vspace{-.2cm}
\subsection{\code{Operation} objects}
\label{sec:operations}
\vspace{-.1cm}

The core abstraction is the \code{Operation} class. Each operation defines:
\begin{tightitemize}
\item A forward \code{apply()} method that maps one representation of the data to another (e.g., from world-frame keypoints to egocentric pose).
\item An \code{invert()} method that reverses the transformation. Many operations of interest are not strictly invertible, e.g.~discretization loses information, thus we only require that \code{op.invert(op.apply(x))} $\approx$ \code{x}. Inverting may also require auxiliary data. 
\item Optional auxiliary data -- output by \code{apply()} and input by \code{invert()} -- captures information needed to recover the original representation. For example, inverting a velocity-encoded representation requires the pose at the start of the sequence. 
\item Dataset-specific parameters (e.g., the bin edges for discretization) that are stored as attributes of the operation. \code{to\_dict()} and \code{from\_dict()} methods serialize these alongside model checkpoints, so the same parameters used during training are reused at inference.
\end{tightitemize}

The library implements a set of general-purpose operations that are useful across many domains -- \code{GlobalVelocity}, \code{LocalVelocity}, \code{Roll}, \code{Zscore}, \code{Discretize}, \code{Fusion}, \code{Subset} -- together with dataset-specific operations that compute biologically motivated representations: \code{Sensory} (with versions for flies and rats) and \code{Pose} (for flies). Adding support for a new domain typically amounts to implementing a new sensory and/or pose operation; the rest of the pipeline composes existing operations.

Most operations act on a single time step. The harder cases -- and the ones that motivate much of the library's design -- are operations that span time, for example velocities. These operations need to interleave correctly with the autoregressive prediction loop, where the model produces one frame at a time and each prediction depends on a temporally consistent input.

\vspace{-.2cm}
\subsection{\code{Data} objects}
\label{sec:data}
\vspace{-.1cm}

Sequence models like transformers operate on temporal chunks of data rather than single frames, and our framework is multi-agent. A \code{Data} object wraps an \code{ndarray} of shape \code{(\ldots, agents, T, d)} together with the list of \code{Operation}s that produced it. Because the chain of operations is stored alongside the data, conversion to the original or an intermediate representation is straightforward.

\vspace{-.2cm}
\subsection{Autoregressive rollout}
\label{sec:rollout}
\vspace{-.1cm}

Rollout proceeds by composing the inverse operations described above with the network's forward pass (\cref{fig:overview}d). At each step, predictions are passed through \code{apply\_inverse\_operations} to recover world-coordinate pose, the world state is updated for all agents, and \code{apply\_operations} is called to produce each agent's sensory input for the next step. Because the forward and inverse pipelines are derived from a shared operation list, switching to a different output representation does not require any change to the rollout machinery: the appropriate inverse chain is constructed automatically from the modified forward chain.

\subsection{\code{Operation} details}
\label{sec:appendix_operation_details}

\begin{figure}
    \centering
    \includegraphics[width=\linewidth]{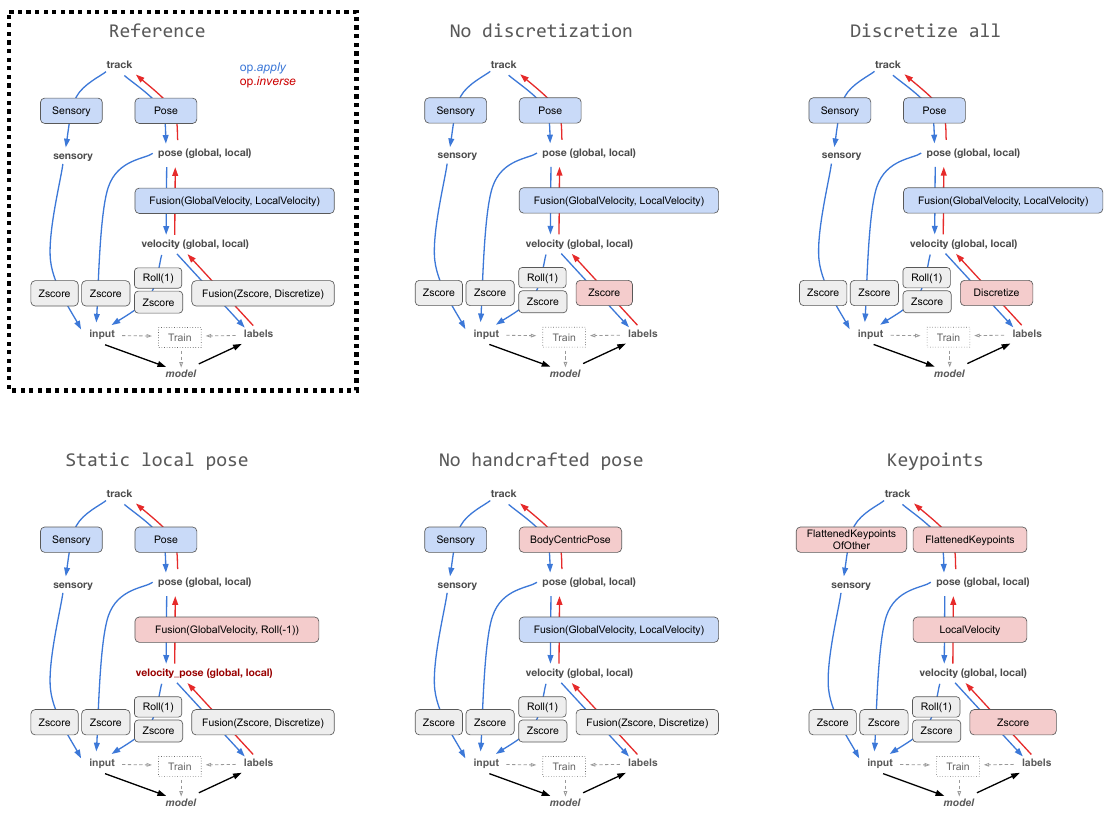}
    \caption{Data flow modifications of variants relative to Reference model.}
    \label{fig:flow_per_variant}
\end{figure}

\begin{figure}
    \centering
    \includegraphics[width=.3\linewidth]{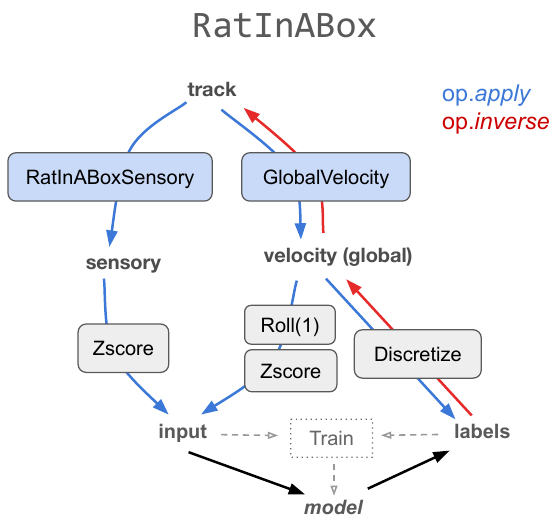}
    \caption{Data flow for RatInABox dataset.}
    \label{fig_new:flow_ratinabox}
\end{figure}

\cref{fig:flow_per_variant} shows the operations involved in computing the input and output representations from the tracked world-frame keypoints of the flies, where the world-frame scale is millimeters and the origin is the center of the arena. \cref{fig_new:flow_ratinabox} shows the operations for the RatInABox model. We describe several of these operations next, for which the implementation is open to interpretation. 

\subsubsection{\code{Sensory} operation (fly-specific)}
\label{sec:appendix_sensory_operation}

The \code{Sensory} operation computes approximation of the agent fly's sensory environment. The input to this operation is the world-frame keypoints, expressed in millimeters in a coordinate frame whose origin is the center of the circular arena. From the perspective of the \emph{agent} fly, \code{Sensory} computes three modes of sensory information (\cref{fig:overview}b): \emph{vision} (the appearance of other flies), \emph{wall touch} (contact with the sloped arena wall), and \emph{other-fly touch} (contact with other flies). These are concatenated into a single sensory vector of length $19 + 72 + 95 = 186$.

\paragraph{Vision.} This mode approximates the agent fly's visual information about other flies in the arena as the minimum distance to other flies in each of 72 directions in the agent fly's coordinate system. We place the viewpoint at the agent's antennae midpoint and define the forward direction $\theta$ as the heading from the base of the thorax to the midpoint of the front thorax. Each other fly is represented by $K_v=4$ keypoints (antennae midpoint, tip of abdomen, and left/right middle-femur bases). For each angular bin, we compute the minimum distance to any other fly. Following \citep{Eyrun2016}, this minimum distance $d$ ($\infty$ if none) is passed through a non-linearity
\begin{equation}
  \rho(d;\,c,\gamma) \;=\; 1 - \min\!\big(1,\; c\, d^{\gamma}\big),
  \label{eq:sensory_nonlinearity}
\end{equation}
We set $\gamma = .6$ and $c=(2R)^{-\gamma}$ with $R=26.689$~mm the arena radius so that the value is $0$ if the distance to the nearest fly is the arena diameter, and $1$ if the nearest fly is touching.

\paragraph{Arena touch.} This mode approximates the tactile information the agent fly has about the height of the arena ceiling, which is the only information the agent fly has about its position in the arena. For each of the 19 keypoints on the agent fly, we return the height of the arena chamber at that keypoint location. As the arena is a dish with a flat floor of radius $R_{\text{in}}=17.5$~mm and ceiling height $H=3.5$~mm, with a rim over which the ceiling slopes to the floor at $R_{\text{out}}=R=26.689$~mm, the arena height is 
\begin{equation}
  w \;=\; \operatorname{clip}\!\Big(H - (r-R_{\text{in}})\tfrac{H}{R_{\text{out}}-R_{\text{in}}},\;0,\;H\Big),
  \qquad w=0 \text{ for } r \ge R_{\text{out}}.
\end{equation}
This is $H$ in the flat center, ramps linearly to $0$ across the rim, and is $0$ outside the arena.

\paragraph{Social touch.} This mode approximates tactile and possibly chemosensory information the agent fly has about what part of other flies it is touching with each of its body parts. Compared to vision, it is only non-zero at very short distances. For each pair of an agent keypoint ($K_t=19$) and an other-fly touch keypoint ($K_o=5$: antennae midpoint, left/right front thorax, base of thorax, tip of abdomen), we compute the minimum distance over other flies between keypoints, resulting in $19 \cdot 5 = 95$ pairs. We pass this distance $d$ through the same nonlinearity $\rho()$ (\cref{eq:sensory_nonlinearity}). Here, the parameters $\gamma = 1.3$ and $c = 0.311$ are set so that the value is $0$ at a distance $2.46 mm$, an estimate of the longest length of any body part. 

\subsubsection{\code{Pose} operation (fly-specific)}
\label{sec:appendix_pose_operation}

The \code{Pose} operation decomposes the 19 tracked world-frame keypoint representation (38-d) into into a compact, interpretable pose vector of $29$ features: $3$ \emph{global} features giving the fly's position and heading in the arena, and $26$ \emph{local} features describing its body configuration in a body-centric frame. 

\paragraph{Global pose.} The keypoints are centered on the midpoint of the two front-thorax keypoints and rotated so that the axis from the base of the thorax to this midpoint points forward. The three global features record this origin ($\code{thorax\_front\_x}$, $\code{thorax\_front\_y}$) and the
heading angle in arena coordinates; all relative features are computed in the rotated frame.

\paragraph{Local pose.}  Each body part is expressed relative to an anatomical pivot as a (distance in mm, angle in rad) pair in polar coordinates, exception: head). The middle legs are represented as kinematic chains: the femur base is placed relative to the thorax, the femur--tibia joint relative to the femur base, and the leg tip relative to the joint, with each angle measured relative to the parent segment so that it encodes a joint angle. Bilaterally paired parts are mirrored across the forward axis on the left side (angle $\alpha \mapsto \pi-\alpha$), so a bilaterally symmetric pose yields identical left- and right-side features. 

\begin{tabular}{llll}
\toprule
Body part & Pivot & Parameterization & \# feats \\
\midrule
Head                       & thorax front      & eye midpoint $(x,y)$ + head angle    & $3$ \\
Abdomen                    & thorax base       & angle                                & $1$ \\
Front leg tip (L/R)        & shoulder          & distance + angle                     & $4$ \\
Middle femur base (L/R)    & mid-thorax        & distance + angle                     & $4$ \\
Middle femur--tibia (L/R)  & femur base        & distance + angle                     & $4$ \\
Middle leg tip (L/R)       & femur--tibia joint & distance + angle                    & $4$ \\
Back leg tip (L/R)         & thorax base       & distance + angle                     & $4$ \\
Wing (L/R)                 & mid-thorax        & angle                                & $2$ \\
\midrule
\multicolumn{3}{l}{Total relative features}                                     & $26$ \\
\bottomrule
\end{tabular}

\textbf{Scale.} Several distances are approximately constant across frames for an individual fly, e.g.~the thorax length and width. These are stored as per-fly scale parameters, and inverting the pose operation thus requires knowing the scale parameters for the specific fly identity. The fly identity to scale parameter mapping is stored with the \code{Pose} operation, and is needed to invert the operation. The per-individual scale comprises $6$ body-part lengths, each computed once per fly as the median over all of that fly's frames. 

\begin{tabular}{ll}
\toprule
Scale parameter & Definition \\
\midrule
Thorax width   & distance between the left and right front-thorax keypoints \\
Thorax length  & distance from the thorax base to the front-thorax midpoint \\
Abdomen length & distance from the thorax base to the tip of the abdomen \\
Wing length    & distance from the mid-thorax to the wing tip (median of both wings) \\
Head width     & distance between the left and right eyes \\
Head height    & distance from the eye midpoint to the antennae midpoint \\
\bottomrule
\end{tabular}

\subsubsection{\code{Discretize} operation (general)}
\label{sec:appendix_discretize_operation}

The \code{Discretize} operation lets model outputs be represented and predicted as probability distributions rather than point estimates, turning regression into classification: each continuous feature is encoded as a categorical distribution over $K$ bins. This lets the model express multimodal, uncertain predictions and be sampled from to generate behavior. We bin features independently into $B=25$ bins.

\textbf{Encoding (soft binning).} A continuous value is not assigned hard to a single bin, but is instead partially assigned to the two adjacent bins whose centers it lies between. Let $c$ and $w$ be the center and width of the bin containing a value $x$, and $d = |x-c|/w \in [0,\tfrac12]$. The value is encoded with mass $1-d$ in that bin and $d$ in the adjacent bin on the side toward $x$: a value at a bin center is one-hot, and a value at a bin edge splits evenly with its neighbor. Mass is never pushed past the first or last bin.

\textbf{Inversion.} Two per-bin statistics are stored when the bins are fit: the median of the training values falling in each bin, and a pool of random training values drawn from each bin. A predicted distribution is converted back to a continuous value either by \emph{sampling} --- draw a bin from the distribution, then a random stored sample from that bin (we store 500 samples per bin), preserving the within-bin distribution --- or \emph{deterministically}, as the bin-median-weighted average of the distribution. Sampling is used when simulating behavior. 

\paragraph{Fitting the bins}

Care must be taken in selecting bin edges as, for many of our features, there are multiple relevant scales. This is particularly acute when modeling velocities. Small differences between two small velocities matter as much as bigger difference between two larger velocities. Previous work has used log or square-root scales of bin sizes to account for this scale issue \citep{Eyrun2016}. For the RatInABox model, we set bin edges based on quantiles of the training data. For the fly behavior model, we set bin edges by optimizing a first-order Markov model, described next. 

\subparagraph{First-order Markov model.}

For the fly behavior pose velocity data, the bin edges for each feature are chosen to maximize the likelihood of its velocity time series under a binned first-order Markov model from the training data. Let $x_t$ be the continuous value at time $t$ and $e_k, \quad k = 0,...,K$ be the bin edges. We introduce a random variable $b_t$ to represent the bin of the value at time $t$, and model 
$$P(x|b_k) = 1/(e_k-e_{k-1}) \text{ if } x \in [e_{k-1},e_k)\text{, } 0 \text{ otherwise}$$
and assume that all inter-timepoint relationships are captured by bin membership: $x_t \perp x_{t-1} | b_t$. Let $b(x)$ be the bin that $x$ falls within. Marginalizing over variables $b_t$ and $b_{t-1}$ yields
$$p(x_t \mid x_{t-1}) = P(x_t | b(x_t)) P(b(x_t) | b(x_{t-1})$$
We empirically estimate the transition probability matrix $P_{ij} = P(b_i | b_j)$ and choose bin edges $\mathbf{e}$ to maximize the resulting log-likelihood under these modeling assumptions:
$$\mathcal{L}(\mathbf{e}) \;=\; \sum_t \left[ \log P_{\,b(x_{t-1}),\,b(x_t)} - \log (e_{b(x_t)} - e_{b(x_t)-1}) \right]$$

The $- \log (e_{b(x_t)} - e_{b(x_t)-1})$ term rewards narrow, high-resolution bins, while the transition term resists over-splitting; the optimum therefore places fine bins where the dynamics are structured and the data dense, and coarse bins elsewhere. The outermost edges are
fixed to the data range. Optionally a pair of edges near the ends is fixed so the extreme bins absorb outliers. 

To optimize the bin edges, edges are initialized with equal-sized bins then optimized by coordinate descent: each edge is moved in turn by a bounded one-dimensional search, with the bin assignments and transition counts updated incrementally, until the log-likelihood converges. 

\subsubsection{\code{GlobalVelocity} operation (general)}
\label{sec:appendix_global_velocity}
The global velocity represents the change in the agent's centroid position and heading in an agent-centric coordinate system, decomposed into the forward, sideways, and angular velocity. Let $(x_t,y_t)$ be the center coordinate and $\theta_t$ the world-frame orientation at time $t$. We project the 1-frame estimate of centroid velocity $(x_{t+1}-x_t,y_{t+1}-y_t)$ onto the coordinate system defined by $\theta_t$ to compute the forward and sideways velocity. Angular velocity is simply $\theta_{t+1}-\theta_t$ wrapped to $[-\pi,\pi]$.

For the fly behavior data, we use the front-thorax midpoint as the centroid and the angle from the base of the thorax to the front-thorax midpoint to define the heading. 

Inverting this operation requires computing the cumulative sum of the velocities over the sequence, and adding it to a stored initial position. 

\subsubsection{\code{LocalVelocity} operation (general)}
\label{sec:appendix_local_velocity}

This operation assumes the inputs are already translated and rotated to an agent-centric coordinate system, and just computes the 1-frame estimate of the velocity of each feature $\mathbf{v}_t = \mathbf{x}_{t+1} - \mathbf{x}_t$. 

As with \code{GlobalVelocity}, inverting this operation requires computing the cumulative sum of the velocities over the sequence, and adding it to a stored initial position. 

\subsubsection{\code{Roll} operation (general)}

It is essential that inputs to the model only be derived from previous frames. With features like velocities depending on multiple frames, we need to shift time series backwards when they are input. The \code{Roll(k)} shifts time series backwards by $k$ frames.

\section{RatInABox data and experiment details}
\label{sec:appendix_ratinabox}

\begin{figure}[htb]
    \centering
    \includegraphics[width=0.7\linewidth]{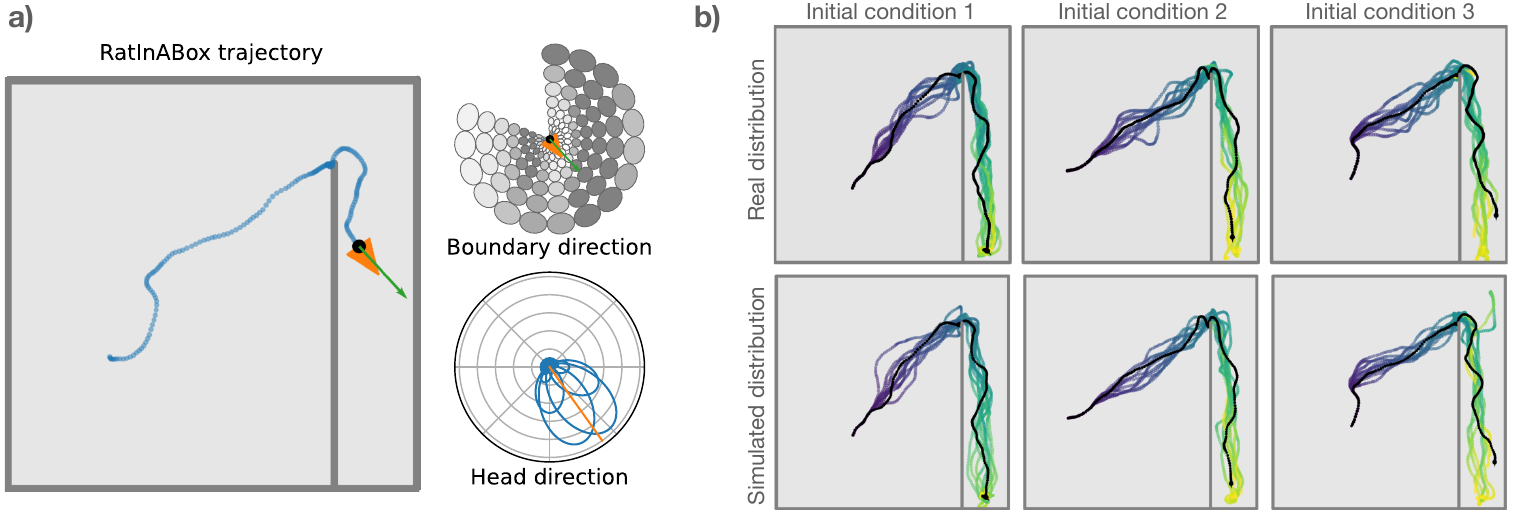}
    \caption{a) RatInABox synthetic trajectory and neural cell activity. b) {\em Top}: Trajectories generated from the "real" RatInABox model from the same initial conditions. One trajectory is highlighted in black, all others are colored by time. {\em Bottom}: Rollouts generated from our trained forecasting model. The black trajectory is the real trajectory highlighted above, colored trajectories are simulations from the trained model. }
    \label{fig:ratinabox_results}
\end{figure}

Above, we demonstrated that our library could be used to build and compare several model variants with different input and output representations. In this Section, we show that our library can easily be applied to other domains. We apply it to synthetic rat behavior generated by RatInABox \citep{RatInABox}, an open-source toolkit for modeling locomotion and neural activity in continuous environments. RatInABox simulates a rodent agent moving through a configurable arena with walls under a physically realistic random motion model fit to the locomotion statistics of real rats. The toolkit also generates synthetic neural activity simulating a variety of cell types. In particular, we used boundary-detection and head-direction neuronal firing rates as the agent's sensory input. Synthetic trajectories were generated with a RatInABox model trained with RL to navigate around a wall to a reward using boundary detection and head direction cells (\cref{fig:ratinabox_results}a, \href{\vidRatInABox}{\suppvid 11}).

We used the \code{AnimalPoseForecasting} library to train a model from these RatInABox trajectories. We used the boundary detection and head direction cell activity (\cref{fig:ratinabox_results}a) as the sensory input, and the discretized egocentric velocity and change in orientation as the output. To do this, we wrote a new \code{Sensory} operation that directly called the RatInABox module code, and otherwise reused existing \code{Operation}s from the \code{AnimalPoseForecasting} library, with a data flow very similar to that for fly trajectories. We trained a transformer model with the same hyperparameters as the fly behavior model. Example rollouts are shown in \cref{fig:ratinabox_results}b. One advantage of testing with synthetic data is we can generate many real trajectories from the same initial conditions, allowing us to compare distributions. The distribution of forecast trajectories looks qualitatively similar to the true distribution from the RatInABox: all trajectories progress around the wall to the goal location with similar variability to the original distribution. 

\subsection{Data generation}
\label{sec:appendix_synthrat_data}

The RatInABox \citep{RatInABox} data are trajectories of a model-free RL agent trained to find a reward hidden behind a wall (\cref{fig:ratinabox_results}a), one of the examples defined in this work, code in the \texttt{reinforcement\_learning\_example} demo notebook. 

To generate the dataset, the trained value function is frozen, and evaluation episodes are rolled out under near-greedy exploitation (exploit/explore ratio $1.0$): $10{,}000$ training episodes and $1{,}000$ validation episodes.

\section{Fly behavior model architecture, training, and simulation}
\label{sec:appendix_fly_model_details}

\subsection{Architecture}
\label{sec:appendix_fly_model_architecture}

All models are transformers that predict the next frame output from a fixed-length history of inputs, where inputs and outputs for each variant are defined by the flow diagrams (\cref{fig:flow_per_variant}). 

\paragraph{Backbone.} The backbone is a transformer with a causal (upper-triangular) attention mask. All models have $10$ layers, $8$ attention heads, embedding dimension $d_{\text{model}}=2048$, feedforward hidden dimension $512$, ReLU activations, and post-layer-norm. Positions are encoded with fixed sinusoidal positional encodings
added to the input embedding (scaled by $\sqrt{d_{\text{model}}}$). The context length is $512$ frames for all variants except the \shorterContextModel variant, which uses $64$ frames context. 

\paragraph{Embedding.}

Each frame's inputs are grouped by modality, embedded to $d_{\text{model}}$ separately, and the per-modality embeddings are summed, with embedding architecture details in the table below. For the majority of model variants, these modalities are each of the sensory modalities: vision, social touch, arena touch, relative pose (approximating proprioception). These are embedded via shallow convolutional networks described in the table below. 1-D networks are convolutional over time, with each dimension treated as a channel. 2-D convolutional networks start with 1 channel, and are convolutional over both time and features. The previous output is fed back autoregressively and embedded with a linear layer. 

\begin{tabular}{lll}
\toprule
Input group & Embedding \\
\midrule
Previous output            & linear \\
Relative pose (proprioception) & 1-D temporal conv ($32/64/128$ channels) \\
Wall touch                 & 1-D temporal conv ($32/64/128$ channels) \\
Vision (other flies)       & 2-D conv ($8/16/32$ channels) \\
Other-fly touch            & 1-D temporal conv ($128/256/512$ channels) \\
\bottomrule
\end{tabular}

\paragraph{Output.} A single linear layer maps each frame's transformer output to a
\emph{continuous} head and a \emph{discrete} head with a soft-max nonlinearity.

\paragraph{Loss.} The prediction loss is a weighted sum of a continuous and a discrete term,
$$\mathcal{L} \;=\; (1-w)\,\mathcal{L}_{\text{cont}} + w\,\mathcal{L}_{\text{disc}},$$
where $\mathcal{L}_{\text{cont}}$ is the $L_1$ (mean absolute) error on the continuous outputs and $\mathcal{L}_{\text{disc}}$ is the cross-entropy of the discrete outputs against their target distributions. The weight $w$ is the fraction of outputs that are discretized. 

\textbf{Optimization.} During training, data are augmented by reflection across the $x$-axis. We use AdamW (learning rate $5\times10^{-5}$, $\beta=(0.9,0.999)$, $\epsilon=10^{-8}$, weight decay $0.01$) with gradient-norm clipping at $1.0$ and a batch size of $64$. The learning rate follows a linear decay from $5\times10^{-5}$ to $0$ over the whole run, with no warmup, stepped every iteration. We train for $500$ epochs, checkpoint every $5$, and keep the checkpoint with the lowest validation loss.

\subsection{Sensory observations}
\label{sec:appendix_synthrat_sensory}

In place of the fly's \code{Sensory} keypoint-based features, the rat model
observes the firing rates of three RatInABox cell populations, computed from each
trajectory using the \code{RatInABox} code library:
\begin{tightitemize}
\item \textbf{Field-of-view boundary cells} (\code{FieldOfViewBVCs}): $128$ boundary-vector cells laid out as an egocentric rect-polar grid of $8$ rings
$\times$ $16$ angles, covering a $\pm150^\circ$ field of view over distances
$0.02$--$0.4$~m. These encode the distance to walls/boundaries in the rat's frame.
\item \textbf{Head-direction cells} (\code{HeadDirectionCells}): $16$ cells
tuned to the head direction.
\item \textbf{Speed cell} (\code{SpeedCell}): $1$ cell encoding movement speed.
\end{tightitemize}
This gives $128+16+1=145$ firing-rate inputs per frame.

\subsection{Pose, velocity, and discretization}
\label{sec:appendix_synthrat_pose}

The rat's pose is its position and head direction; there is no articulated
(relative) pose. The prediction target is the global velocity computed by
\code{GlobalVelocity} --- the frame-to-frame \emph{forward}, \emph{sideways}, and
\emph{orientation} (heading) change, in the rat's own heading frame
(\cref{sec:appendix_global_velocity}). All three global-velocity components are discretized into $K=25$ bins each. Unlike the fly model, the rat uses simple equal-frequency (quantile) bin edges fit from the training data at train time.

\subsection{Model architecture and training}
\label{sec:appendix_synthrat_model}

The model architecture and training match the fly behavior model architecture, with the following differences. Context is $16$ frames, and the model is trained for 20,000 epochs. 

\paragraph{Embedding.} Each sensory population is embedded separately by a
small convolutional network before entering the transformer: a 2-D convolution over
the $8\times16$ boundary-cell grid (channel widths $32{\to}64{\to}128$), and 1-D
temporal convolutions for the head-direction cells ($16{\to}32{\to}64$) and the
speed cell ($16{\to}32{\to}64$).

\section{Simulation rollouts}
\label{sec:appendix_rollouts}

To evaluate a trained model we simulate behavior from it and compare the simulated trajectories to real trajectories. This section describes the autoregressive rollout procedure. A rollout simulates a subset of the agents. For the fly behavior dataset, all males are simulated, while the females keypoints are the real flies' trajectories. The first \code{contextl} frames are copied from real data (the prompt); thereafter the flies' behavior is generated one frame at a time for $512$ further frames. At each step, the transformer is run on the current context (a causal sliding window of \code{contextl} frames), and predicts the next frame's output. By inverting the data flow \code{Operation}s, we compute the world-frame keypoints. This is done independently for all simulated agents. Rollouts for the simulated agents are closed-loop: after each step the sensory features are recomputed from the combined keypoints of all agents and the real animals. Within a rollout, the group of simulated agents are thus simulated jointly and interact while any other animals are held at their real trajectories, and the simulated agents still sense them. Flies can therefore approach, chase, and avoid one another purely through this sensory feedback loop.

Discretized outputs are sampled stochastically: a bin is drawn from its softmax distribution, and a continuous value is then drawn uniformly from the pool of real training values that fell in that bin (\cref{sec:appendix_discretize_operation}). Continuous outputs are taken deterministically, as the network's real-valued prediction.

We index each generated frame by its distance to the prompt (DTP), the number of frames since the end of the burn-in ($1$ for the first generated frame, up to $512$). Metrics are reported over the first $64$ or all $512$ generated frames, or plotted directly against DTP.

Rollouts are seeded from the training trajectories. Every sufficiently long multi-fly window is simulated, strided by $512$ frames. The examples in the Supplementary videos use a set of $11$ evenly spaced sessions.

\end{document}